\documentclass[aip,jcp,preprint,superscriptaddress]{revtex4-1}
\usepackage[dvips]{graphicx}
\usepackage{epsfig}
\usepackage{bm}
\usepackage{psfrag}
\usepackage{xspace}
\usepackage[percent]{overpic}
\usepackage[utf8]{inputenc}
\usepackage{amsmath}
\usepackage{booktabs}

\usepackage{microtype}
\usepackage{color}
\usepackage{subcaption}
\usepackage{epsf}

\usepackage{amssymb}

\usepackage{bmpsize}

\usepackage{mathtools}
\newcommand{\vect}[1]{\mathbf{#1}}

\newcommand{\fsz}{\footnotesize}

\newcommand{\vq}{{\boldsymbol q}}

\newcommand{\jmst}{J. Mol. Struct.}
\begin{document}

\title{Perspective: Energy Landscapes for Machine Learning}

\author{Andrew J. Ballard}
\affiliation{University Chemical Laboratories, Lensfield Road, Cambridge CB2 1EW, United Kingdom}
\author{Ritankar Das}
\affiliation{University Chemical Laboratories, Lensfield Road, Cambridge CB2 1EW, United Kingdom}
\author{Stefano Martiniani}
\affiliation{University Chemical Laboratories, Lensfield Road, Cambridge CB2 1EW, United Kingdom}
\author{Dhagash Mehta}
\affiliation{Department of Applied and Computational Mathematics and Statistics, University of Notre Dame, IN, USA}
\author{Levent Sagun}
\affiliation{Mathematics Department, Courant Institute, New York University, NY, USA}
\author{Jacob D. Stevenson}
\affiliation{Microsoft Research Ltd, 21 Station Road, Cambridge, CB1 2FB, UK}
\author{David J. Wales}
\email[]{dw34@cam.ac.uk}
\affiliation{University Chemical Laboratories, Lensfield Road, Cambridge CB2 1EW, United Kingdom}

\begin{abstract}
Machine learning techniques are being increasingly used as flexible
non-linear fitting and prediction tools in the physical sciences.
Fitting functions that exhibit multiple solutions as local minima
can be analysed in terms of the corresponding machine learning landscape.
Methods to explore and visualise molecular potential energy landscapes
can be applied to these machine learning landscapes to gain new insight into
the solution space involved in training and the nature of the corresponding predictions.
In particular, we can define quantities analogous to molecular structure, thermodynamics,
and kinetics, and relate these emergent properties to the structure of the underlying landscape.
This Perspective aims to describe these analogies with examples from 
recent applications, and suggest avenues for new interdisciplinary research.
\end{abstract}

\maketitle

\section{Introduction}

Optimisation problems abound in computational science and technology. From
force field development to thermodynamic sampling, bioinformatics and
computational biology, optimisation methods are a crucial ingredient of most
scientific disciplines \cite{ChillSRSXFWH14}. 
Geometry optimisation is a key component of the
potential energy landscapes approach in molecular science, where emergent
properties are predicted from local minima and the transition states that connect
them \cite{Wales03,Wales10a}. This formalism has been applied to a wide variety of physical systems, including
atomic and molecular clusters, biomolecules, mesoscopic models, and glasses, to understand their
structural properties, thermodynamics and kinetics. The methods involved in the
computational potential energy landscapes approach amount to optimisation of a
non-linear function (energy) in a high-dimensional space (configuration space).
Machine learning problems employ similar concepts: the training of a model is
performed by optimising a cost function with respect to a set of adjustable parameters. 

Understanding how emergent observable properties of molecules and condensed matter
are encoded in the underlying potential energy surface is a key motivation in
developing the theoretical and computational aspects of energy landscape research.
The fundamental insight that results has helped to unify our understanding of
how structure-seeking systems, such as `magic number' clusters, functional biomolecules,
crystals and self-assembling structures, differ from amorphous materials and
landscapes that exhibit broken ergodicity \cite{walesmw98,Wales05,deSouzaW08,Wales10a}.
Rational design of new molecules and materials can exploit this insight, for example
associating landscapes for self-assembly with a well defined free energy minimum
that is kinetically accessible over a wide range of temperature.
This structure, where there are no competing morphologies separated from the target by high barriers,
corresponds to an `unfrustrated' landscape \cite{bryngelsonosw95,onuchicwls97,Wales03,Wales05}.
In contrast, designs for multifunctional systems, including molecules with the capacity to
act as switches, correspond to multifunnel landscapes \cite{ChakrabartiW11,ChebaroBCW15}.

In this Perspective we illustrate how the principles and tools of the potential energy landscape 
approach can be applied to machine learning (ML) landscapes.
Some initial results are presented, which indicate how this view 
may yield new insight into ML training and prediction in the future.
We hope our results will be interesting to both the energy landscapes and
machine learning communities. In particular, it is of fundamental interest
to compare these ML landscapes to those that arise for molecules and condensed matter.
The ML landscape provides both a means to visualise and
interpret the cost function solution space and a computational framework for
quantitative comparison of solutions.

\section{The Energy Landscape Perspective}
\label{sec:energy-landscapes}

The potential energy function in molecular science is a surface defined in a
(possibly very high-dimensional) configuration space, which represents all
possible atomic configurations \cite{mezey87,Wales03}.
In the potential energy landscape approach,
this surface is divided into basins of attraction, each defined by the steepest-descent 
pathways that lead to a particular local minimum \cite{mezey87,Wales03}.
The mapping from a continuous multidimensional surface to local minima can be very
useful. In particular, it provides a route to prediction of structure and
thermodynamics \cite{Wales03}.
Similarly, transitions between
basins can be characterised by geometric transition states (saddle points of index
one), which lie on the boundary between one basin and
another \cite{Wales03}.
Including these transition states in our description of
the landscape produces a kinetic transition network \cite{NoeF08,pradag09,Wales10a},
and access to dynamical properties and
`rare' events \cite{Wales02,dellagobc98,PasseroneP01,ERV02}.
The pathways mediated by these transition states correspond to processes such as molecular
rearrangements, or atomic migration.
For an ML landscape we can define the connectivity between minima that represent 
different locally optimal fits to training data in an analogous fashion.
To the best of our knowledge,
interpreting the analogue of a molecular rearrangement mechanism 
for the ML landscape has yet to be explored.

Construction of a kinetic transition network \cite{NoeF08,pradag09,Wales10a} also provides a
convenient means to visualise a high-dimensional surface. Disconnectivity
graphs \cite{beckerk97,walesmw98} represent the landscape in terms of 
local minima and connected transition states, reflecting the barriers
and topology through basin connectivity. The overall
structure of the disconnectivity graph can provide immediate insight into 
observable properties \cite{walesmw98}: a single-funnelled landscape typically
represents a structure-seeking system that equilibrates rapidly, whereas multiple funnels indicate
competing structures or morphologies, which may be
manifested as phase transitions and even glassy phenomenology. Locating the
global minimum is typically much easier for single funnel landscapes \cite{doyemw99a}. 

The decomposition of a surface into minima and transition states is
quite general and can naturally be applied to systems that do not
correspond to an underlying molecular model. In particular, we can use this strategy for
machine learning applications, where training a model amounts to
minimisation of a cost function with respect to a set of parameters. In the
language of energy landscapes, the machine learning cost function plays the
role of energy, and the model parameters are the `coordinates' of the
landscape. The minimised structures represent the optimised model parameters
for different training iterations. The transition states are the 
index one saddle points of the landscape \cite{murrelll68}.

Energy landscape methods \cite{Wales03} could be particularly beneficial to the
ML community, where non-convex optimisation has sometimes been viewed as
less appealing, despite supporting richer models with superior
scalability \cite{Collobert2006}.
The techniques described below could provide
a useful computational framework for exploring and visualising ML landscapes,
and at the very least, an alternative view to non-convex optimisation.
The first steps in this direction have recently been reported \cite{PavlovskaiaTZ14,BallardSDW16,DasW16}.
The results may prove to be useful for various applications of machine learning in
the physical sciences. 
Examples include fitting potential energy surfaces,
where neural networks have been used extensively \cite{1.2336223,Houlding2007,PhysRevLett.100.185501,B905748J,1.2746232,1.4936660,1.4961454,doi:10.1021/jp9105585}
for at least 20 years  \cite{1.469597,browngc96,doi:10.1021/jp972209d}. 
Recent work includes
prediction of binding affinities in protein-ligand complexes \cite{Ballester01052010},
applications to the design of novel materials \cite{doi:10.1021/jp500350b,1.4962754},
and refinement of transition states \cite{PozunHSRMH12} using support vector machines \cite{Cortes1995}.

In the present contribution we illustrate the use of techniques from the energy landscapes field
to several ML examples, including non-linear
regression, and neural network classification. When surveying the cost function
landscapes, we employed the same techniques and algorithms as for 
the molecular and condensed matter systems of interest in the physical sciences:
specifically, local and global minima were obtained with the basin-hopping
method \cite{lis87,lis88,WalesD97} using a customised LBFGS minimisation
algorithm \cite{Nocedal80}. Transition state searches
employed the doubly-nudged \cite{TrygubenkoW04,TrygubenkoW04b} elastic band \cite{HenkelmanUJ00,HenkelmanJ00}
approach and hybrid eigenvector-following \cite{munrow99,ZengXH14}.
These methods are all well established, and will not be described in detail here.
We used the python-based energy landscape explorer pele \cite{pele}, with a customised
interface for ML systems, 
along with the {\tt GMIN} \cite{gmin},
{\tt OPTIM} \cite{optim},  and {\tt PATHSAMPLE} \cite{pathsample} programs, available for download under the
Gnu General Public Licence.


\section{Prediction for Classification of Outcomes in Local Minimisation}
\label{sec:LJAT3}

Neural networks (NN) have been employed in two previous classification problems that 
analyse the underlying ML landscape, namely predicting which isomer results from a molecular geometry
optimisation \cite{BallardSDW16} and for patient outcomes in a medical diagnostic context \cite{DasW16}.
Some of the results from the former study will be illustrated here,
and we must carefully distinguish isomers corresponding to minima of a molecular potential
energy landscape from the ML landscape of solutions involved in predicting which of
the isomers will result from geometry optimisation starting from a given molecular configuration.
We must also distinguish this classification problem from ab initio structure prediction: the possible
outcomes of the geometry optimisation must be known in advance, either in terms of distinct isomers,
or the range that they span in terms of potential energy or appropriate structural order parameters
for larger systems.
The ability to make predictions that are sufficiently reliable could produce significant savings
in computational effort for applications that require repeated local minimisation.
Examples include basin-sampling for calculating global thermodynamic properties in systems subject
to broken ergodicity \cite{Wales13},
contruction of kinetic transition networks \cite{NoeKSF06},
and methods to estimate the volume of basins of attraction for jammed packings, 
which provide measures of configurational entropy in granular packings.\cite{MartinianiSSWF16}
Here the objective would be to terminate the local minimisation 
as soon as the outcome could be predicted with sufficient confidence to 
converge the properties of interest \cite{SwerskySA14,BallardSDW16}.

The test system considered here is a simple triatomic cluster with four distinguishable local minima
that represent the possible outcomes for local minimisation.
This system has previously served as a benchmark for visualising the performance of different
geometry optimisation approaches \cite{Wales92,Wales93d,AsenjoSWF13}.
The potential energy, $V$, is a sum of pairwise terms, corresponding to the Lennard-Jones form \cite{jonesi25},
and the three-body Axilrod--Teller function \cite{AxilrodT43}, which represents an instantaneous induced dipole-induced dipole interaction:
\begin{eqnarray}
\label{eq:pot}
V&=&4\varepsilon \sum_{i<j}\left[\left(\frac{\sigma}{r_{ij}}\right)^{12}
- \left(\frac{\sigma}{r_{ij}}\right)^6 \right] +
\gamma\sum_{i<j<k}\left[
\frac{1+3\cos\theta_1\cos\theta_2\cos\theta_3}{(r_{ij}r_{ik}r_{jk})^3} \right],
\end{eqnarray}
where  $\theta_1$, $\theta_2$ and $\theta_3$ are the internal angles of
the triangle formed by atoms $i$, $j$, $k$.
$r_{ij}$ is the distance between atoms $i$ and $j$.
The influence of the three-body term is determined by the magnitude of the parameter $\gamma$, and
we use $\gamma=2$, where the equilateral triangle ($V=-2.185\,\varepsilon$) competes with
three permutational isomers of a linear minimum ($V=-2.219\,\varepsilon$).
In the triangle the bond length is 1.16875$\,\sigma$, and in the linear minima
the distance from the centre atom to its neighbours is 1.10876$\,\sigma$.

The objective of our machine learning calculations for this system is a classification,
to predict which of the four minima a local minimisation would converge to, given 
data for one or more configurations. 
The data in question could be combinations of the energy, gradient, and geometrical parameters
for the structures in the optimisation sequence.
Our initial tests, which are mostly concerned with the structure of the ML solution landscape,
employed the three interparticle separations $r_{12}$, $r_{13}$ and $r_{23}$ as data \cite{BallardSDW16}.
Inputs were considered for separations corresponding to the initial geometry,
and for one or more configurations in the minimisation sequence.

A database of 10,000 energy minimisations, each initiated from different atomic coordinates 
distributed in a cube of side length $2\sqrt{3}\,\sigma$, 
was created using the customised LBFGS routine in our {\tt OPTIM} program \cite{optim} (this is
a limited memory quasi-Newton Broyden,\cite{Broyden70}
Fletcher,\cite{Fletcher70} Goldfarb,\cite{Goldfarb70} Shanno,\cite{Shanno70} scheme).
Each minimisation was converged until the root mean square gradient fell below $10^{-6}$ reduced units,
and the outcome (one of the four minima) was recorded.

\begin{figure}[htp]
\begin{centering}
\psfrag{inputs}[bc][bc]{inputs}
\psfrag{outputs}[bc][bc]{outputs}
\psfrag{hidden layer}[tc][tc]{hidden layer}
\psfrag{w2jk}[tc][tc]{$w^{(2)}_{jk}$}
\psfrag{w1ij}[tc][tc]{$w^{(1)}_{ij}$}
\psfrag{bias}[bc][bc]{bias}
\psfrag{bh}[bc][bc]{$w^{\rm bh}_j$}
\psfrag{bo}[bc][bc]{$w^{\rm bo}_i$}
\psfrag{x1}[cr][cr]{$x_1$}
\psfrag{x2}[cr][cr]{$x_2$}
\psfrag{x3}[cr][cr]{$x_3$}
\psfrag{x4}[cr][cr]{$x_4$}
\psfrag{y1}[cr][cr]{$y_1$}
\psfrag{y2}[cr][cr]{$y_2$}
\psfrag{y3}[cr][cr]{$y_3$}
\psfrag{y4}[cr][cr]{$y_4$}
\includegraphics[width=0.9\textwidth]{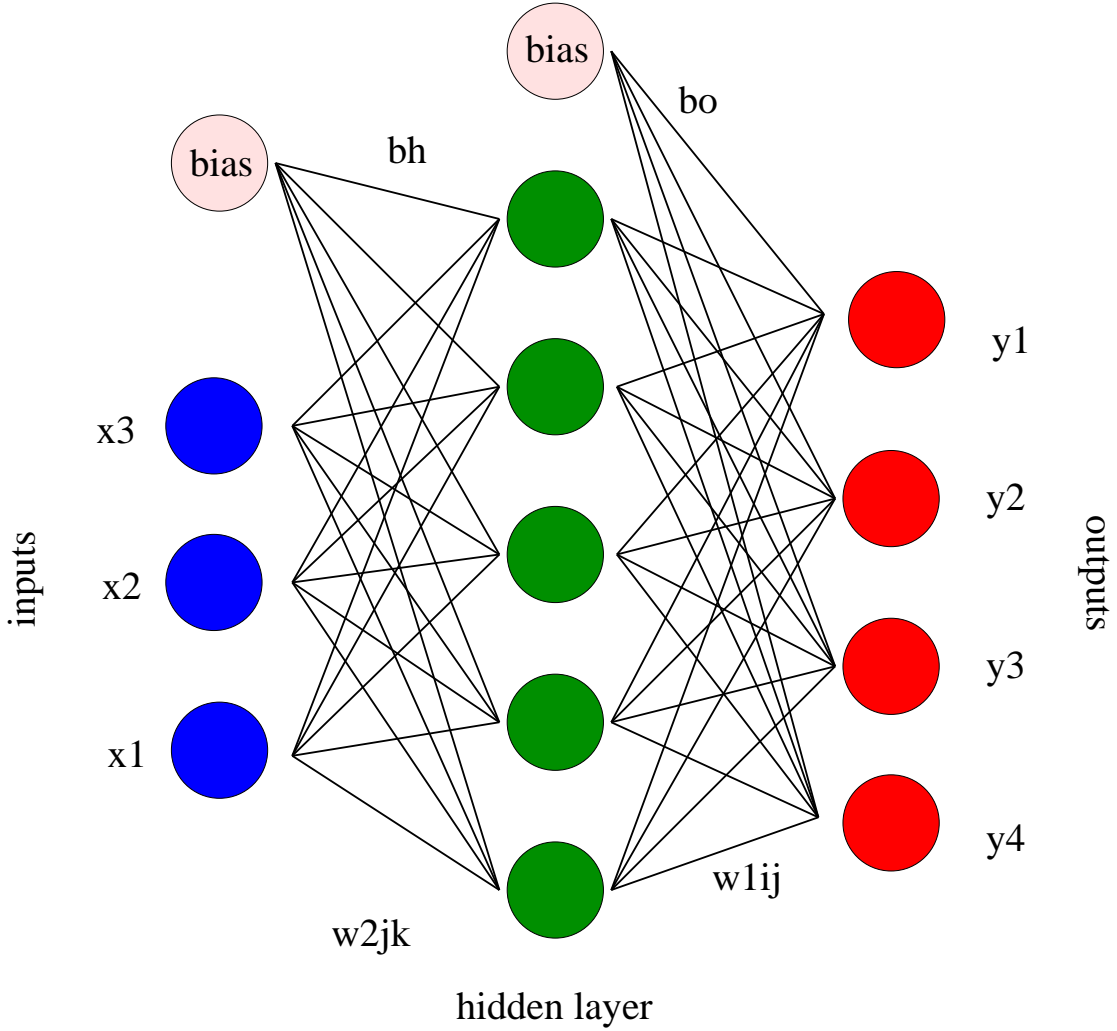}
\caption{Organisation of a three-layer neural network
with three inputs, five hidden nodes, and four outputs.
The training variables are the link weights, $w^{(2)}_{jk}$, $w^{(1)}_{ij}$,
and the bias weights, $w^{\rm bh}_j$ and $w^{\rm bo}_i$.
}
\label{fig:MLP}
\end{centering}
\end{figure}

The neural networks used in the applications discussed below all have three layers,
corresponding to input, output, and hidden nodes \cite{Bishop06}.
A single hidden layer has been used successfully in a variety of previous applications \cite{HastieTF09}.
A bias was added to the sum of weights used in the activation function for each 
hidden node, $w^{\rm bh}_j$, and each output node, $w^{\rm bo}_i$,
as illustrated in Figure \ref{fig:MLP}.
For inputs we consider ${\bf X}=\{{\bf x}^1,\ldots,{\bf x}^{N_{\rm data}}\}$, 
where $N_{\rm data}$ is the number of minimisation sequences in the training or test set,
each of which has dimension $N_{\rm in}$,
so that ${\bf x}^\alpha = \{x^\alpha_1,\ldots,x^\alpha_{N_{\rm in}}\}$.
For this example
there are four outputs corresponding to the four possible local minima (as in Figure \ref{fig:MLP}),
denoted by $y^{\rm NN}_i$, with
\begin{equation}
y^{\rm NN}_i = w^{\rm bo}_i + \sum_{j=1}^{N_{\rm hidden}} w^{(1)}_{ij}
      \tanh\left[w^{\rm bh}_j + \sum_{k=1}^{N_{\rm in}} w^{(2)}_{jk}x_k  \right],
\label{eq:yNN}
\end{equation}
for a given input {\bf x} and link weights
$w^{(1)}_{ij}$ between hidden node $j$ and output $i$, and $w^{(2)}_{jk}$ between
input $k$ and hidden node $j$.

The four outputs were converted into softmax probabilities as
\begin{equation}
\label{eq:softmax}
p^{\rm NN}_{c}({\bf W};{\bf X}) = e^{y^{\rm NN}_{c}}/ \sum_{a=0}^{3} e^{y^{\rm NN}_{a}}.
\end{equation}
This formulation is designed to reduce the effect of outliers.

In each training phase we minimise the cost (objective) function, $E^{\rm NN}({\bf W};{\bf X})$,
with respect to the variables $w^{(1)}_{ij}$, $w^{(2)}_{jk}$, $w^{\rm bh}_j$ and $w^{\rm bo}_i$, which are
collected into the vector of weights ${\bf W}$.
An $L^2$ regularisation term is included in $E^{\rm NN}({\bf W};{\bf X})$,
corresponding to a sum of squares of the independent variables, 
multiplied by a constant coefficient $\lambda$, which is chosen in advance and fixed.
This term is intended to prevent overfitting; it disfavours large values for individual variables.
We have considered applications where the regularisation is applied to all the variables,
and compared the results with a sum that excludes the bias weights.
Regularising over all the variables has the advantage of eliminating the zero Hessian eigenvalue
that otherwise results from an additive shift in all the $w^{\rm bo}_i$.
Such zero eigenvalues are a consequence of continuous symmetries in the cost function (Noether's theorem).
For molecular systems such modes commonly arise from overall translation and rotation, and
are routinely dealt with by eigenvalue shifting or projection using the known analytical forms for the
eigenvectors \cite{pagem88,Wales03}.
Larger values of the parameter $\lambda$ simplify the landscape, reducing the number of minima. This
result corresponds directly with the effect of compression for a molecular system \cite{PhysRevE.62.8753},
which has been exploited to accelerate global optimisation.
A related hypersurface deformation approach has been used to treat graph partitioning problems \cite{Stillinger1988}.

For each LBFGS geometry optimisation sequence, $d$, with $N_{\rm in}$ inputs collected into data item ${\bf x}^d$,
we know the actual outcome or class label, $c(d)=0$, 1, 2 or 3, corresponding to the local minimum at convergence.
The networks were trained using either 500 or 5,000 of the LBFGS sequences, chosen at random with no overlap,
by minimising
\begin{equation}
\label{eq:obj}
E^{\rm NN}({\bf W};{\bf X}) = -\frac{1}{N_{\rm data}}\sum_{d=1}^{N_{\rm data}} \ln p^{\rm NN}_{c(d)}({\bf W};{\bf X}) +
    \lambda {\bf W}^2.
\end{equation}
Results were compared for different values of $\lambda$ and in some cases for regularisation excluding the bias weights.
These formulations, including analytic first and second derivatives with respect to ${\bf W}$,
have been programmed in our {\tt GMIN} global
optimisation program \cite{gmin}
and in our {\tt OPTIM} code for analysing stationary points and pathways \cite{optim}.
$E^{\rm NN}({\bf W};{\bf X})$ was minimised using the same customised LBFGS routine that
was employed to create the database of minimisation sequences for the triatomic cluster.

In the testing phase the variables ${\bf W}$ are fixed for a particular local minimum of
$E^{\rm NN}({\bf W};{\bf X}_{\rm train})$ obtained with the training data, and we evaluate
$E^{\rm NN}({\bf W};{\bf X}_{\rm test})$ for 500 or 5,000 of the minimisation sequences outside the training set.
The results did not change significantly between the larger and smaller training and testing sets.

\begin{figure}[htp]
 \begin{centering}
\psfrag{a}[cc][cc]{(a)}
\psfrag{b}[cc][cc]{(b)}
\psfrag{c}[cc][cc]{(c)}
\psfrag{d}[cc][cc]{(d)}
\psfrag{Order Parameter}[cc][cc]{}
\psfrag{epsilon}[cl][cl]{\fsz 20}
\psfrag{    0.40 }[cl][cl]{\fsz 0.4}
\psfrag{    0.30 }[cl][cl]{\fsz 0.3}
\psfrag{    0.20 }[cl][cl]{\fsz 0.2}
\psfrag{    0.10 }[cl][cl]{\fsz 0.1}
\psfrag{    0.00 }[cl][cl]{\fsz 0.0}
\psfrag{    0.00 }[cl][cl]{\fsz 0.00}
\psfrag{    0.04 }[cl][cl]{\fsz 0.04}
\psfrag{    0.07 }[cl][cl]{\fsz 0.07}
\psfrag{    0.11 }[cl][cl]{\fsz 0.11}
\psfrag{    0.14 }[cl][cl]{\fsz 0.14}
\includegraphics[width=0.875\textwidth]{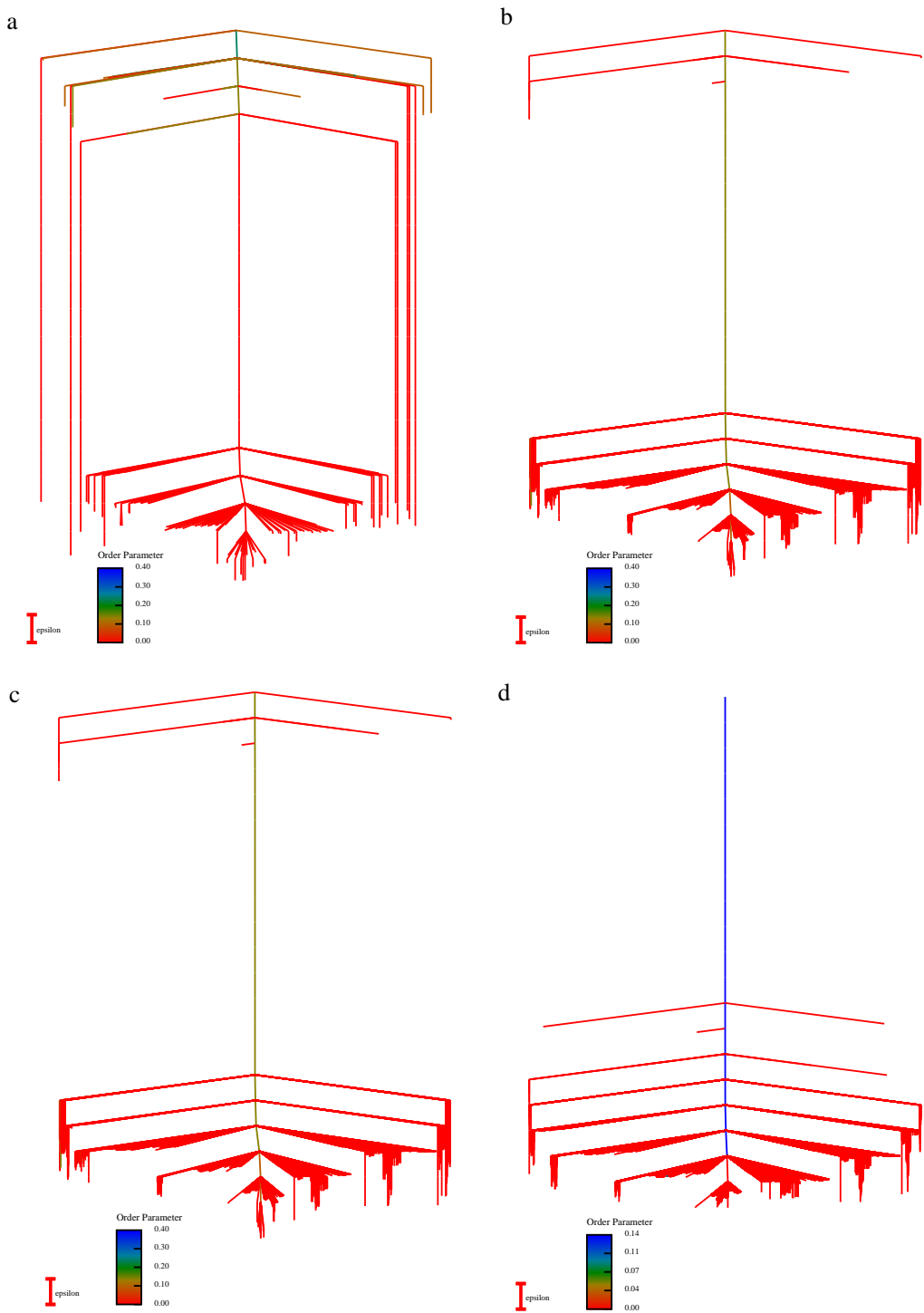}
\caption{Disconnectivity graphs for the fitting landscapes
of a triatomic cluster. Three inputs were used
for each minimisation sequence, corresponding to 
the three interatomic distances in the initial configuration. 
These graphs are for neural networks with 
(a) 3, (b) 4, (c) 5, and (d) 6 hidden nodes.
The branches are coloured according to the misclassification distance for the
local minima evaluated using training data, as described in \S \ref{sec:NN}.
}
  \label{fig:one_time}
  \end{centering}
  \end{figure}

We first summarise some results for ML landscapes corresponding to input data for the three
interparticle distances at each initial random starting point.
The number of local minima obtained \cite{BallardSDW16} was 162, 2,559, 4,752 and 19,045 for
three, four, five and six hidden nodes, respectively, with
1,504, 10,112, 18,779 and 34,052
transition states.
The four disconnectivity graphs are shown in Figure \ref {fig:one_time}.
In each case the vertical axis corresponds to $E^{\rm NN}({\bf W};{\bf X}_{\rm train})$, and branches terminate at the values for
particular local minima.
At a regular series of thresholds for $E^{\rm NN}$ we perform a superbasin analysis \cite{beckerk97},
which segregates the ML solutions into disjoint sets.
Local minima that can interconvert via a pathway where the highest transition state lies below the threshold are in the same superbasin.
The branches corresponding to different sets or individual minima merge at the threshold energy where they
first belong to a common superbasin.
In this case we have coloured the branches according to the misclassification index, discussed
further in \S \ref{sec:NN}, which is defined as the fraction of
test set images that are misclassified by the minimum in question or the global minimum, but not both.
All the low-lying minima exhibit small values,
meaning that they perform much like the global minimum. The index
rises to between 0.2 and 0.4 for local minima with higher values of
$E^{\rm NN}({\bf W};{\bf X}_{\rm train})$.
These calculations were performed using the  pele \cite{pele} ML interface
for the formulation in Eq.~(\ref{eq:yNN}), where regularisation excluded
the weights for the bias nodes \cite{BallardSDW16}.

When more hidden nodes are included the dimensionality of the ML landscape increases,
along with the number of local minima and transition states. 
This observation is in line with well known results for molecular systems: as the
number of atoms and configurational degrees of freedom increases the number of minima
and transition states increases exponentially \cite{stillingerw84,WalesD03}.
The residual error reported by $E^{\rm NN}({\bf W};{\bf X}_{\rm train})$ decreases 
as more parameters are included, and so there is a trade-off between the complexity of
the landscape and the quality of the fit \cite{BallardSDW16}.

The opportunities for exploiting tools from the potential energy landscape framework 
have yet to be investigated systematically.
As an indication of the properties that might be of interest we now illustrate a
thermodynamic analogue corresponding to the heat capacity, $C_V$.
Peaks in $C_V$ are particularly insightful, and in molecular systems they correspond
to phase-like transitions.
Around the transition temperature the occupation probability shifts between local minima
with qualitatively different properties in terms of energy and entropy:
larger peaks correspond to greater differences \cite{walesd95}.
For the ML landscape we would interpret such features in terms of fitting solutions with
different properties.
Understanding how and why the fits differ could be useful in combining solutions to
produce better predictions. Here we 
simply illustrate some representative results, which suggest that ML landscapes may support
an even wider range of behaviour than molecular systems.

To calculate the $C_V$ analogue
we use the superposition approach \cite{stillingerw84,wales93f,Wales03,stillinger95,StrodelW08b,SharapovMM07}
where the total partition function, $Z(T)$, 
is written as a sum over the partition functions $Z_\alpha(T)$, for all the local minima, $\alpha$.
This formulation can be formally exact, but is usually applied in the harmonic approximation 
using normal mode analysis to represent the vibrational density of states.
The normal mode frequencies are then related to the eigenvalues of the Hessian
(second derivative) matrix, $\mu_{\alpha}(i)$, for local minimum $\alpha$:
\begin{equation}
Z(T)=\sum_\alpha Z_\alpha(T) \approx
\sum_\alpha \frac{e^{-\beta E_{\alpha}}}{(\beta/2\pi)^\kappa \prod_{i=1}^\kappa \mu_{\alpha}(i)^{1/2}},
\label{eq:z}
\end{equation}
Here $\kappa$ is the number of non-zero eigenvalues,
$\beta=1/k_BT$, $k_B$ is the Boltzmann constant,
and $E^{\rm NN}({\bf W}_{\alpha};{\bf X}_{\rm train})$ is the objective (loss) function for minimum $\alpha$.
$T$ plays the role of temperature in this picture, with $C_V(T)=(1/k_BT^2) \partial^2 \ln Z(T)/\partial \beta^2$.

Many molecular and condensed matter systems exhibit a $C_V$ peak corresponding to a first order-like melting
transition, where the occupation probability shifts from a relatively small number of low energy minima,
to a much larger, entropically favoured, set of higher energy structures, which are often more disordered.
However, low temperature peaks below the melting temperature, corresponding to solid-solid transitions, can also occur.
These features are particularly interesting, because they suggest the presence of competing low energy
morphologies, which may represent a challenge for structure prediction \cite{OakleyJW13}, and lead
to broken ergodicity \cite{NeirottiCFD00,CalvoNFD00,MandelshtamFC06,SharapovMM07,SharapovM07,Calvo10,Wales13,SehgalMF14},
and slow interconversion rates that constitute `rare events' \cite{Wales02,Wales04,PiccianiAKT11}.
Initial surveys of the $C_V$ analogue for ML landscapes, calculated using analytical second derivatives
of $E^{\rm NN}({\bf W}_{\alpha};{\bf X})$, produced plots with multiple peaks,
suggesting richer behaviour than for molecular systems.

\begin{figure}[htp]
\psfrag{ 13.5}[cr][cr]{\fsz 13.5}
\psfrag{ 13.6}[cr][cr]{\fsz 13.6}
\psfrag{ 13.7}[cr][cr]{\fsz 13.7}
\psfrag{ 13.8}[cr][cr]{\fsz 13.8}
\psfrag{  15}[cr][cr]{\fsz 15}
\psfrag{ 17}[cr][cr]{\fsz 17}
\psfrag{ 19}[cr][cr]{\fsz 19}
\psfrag{ 21}[cr][cr]{\fsz 21}
\psfrag{0.0}[tc][tc]{\fsz 0.0}
\psfrag{0.4}[tc][tc]{\fsz 0.4}
\psfrag{0.8}[tc][tc]{\fsz 0.8}
\psfrag{1.2}[tc][tc]{\fsz 1.2}
\psfrag{1.6}[tc][tc]{\fsz 1.6}
\psfrag{2.0}[tc][tc]{\fsz 2.0}
\psfrag{ 0}[tc][tc]{\fsz 0}
\psfrag{ 4}[tc][tc]{\fsz 4}
\psfrag{ 8}[tc][tc]{\fsz 8}
\psfrag{ 12}[tc][tc]{\fsz 12}
\psfrag{ 10}[cr][cr]{10}
\psfrag{ 20}[cr][cr]{20}
\psfrag{ 30}[cr][cr]{30}
\psfrag{ 40}[cr][cr]{40}
\psfrag{ 50}[cr][cr]{50}
\psfrag{ 60}[cr][cr]{60}
\psfrag{ 70}[cr][cr]{70}
\psfrag{0}[tc][tc]{0}
\psfrag{5}[tc][tc]{5}
\psfrag{10}[tc][tc]{10}
\psfrag{15}[tc][tc]{15}
\psfrag{20}[tc][tc]{20}
\psfrag{25}[tc][tc]{25}
\psfrag{30}[tc][tc]{30}
\psfrag{124}[tr][tr]{124}
\psfrag{152}[bl][bl]{152}
\psfrag{2}[tl][tl]{2}
\centerline{
 \includegraphics[width=1.0\textwidth]{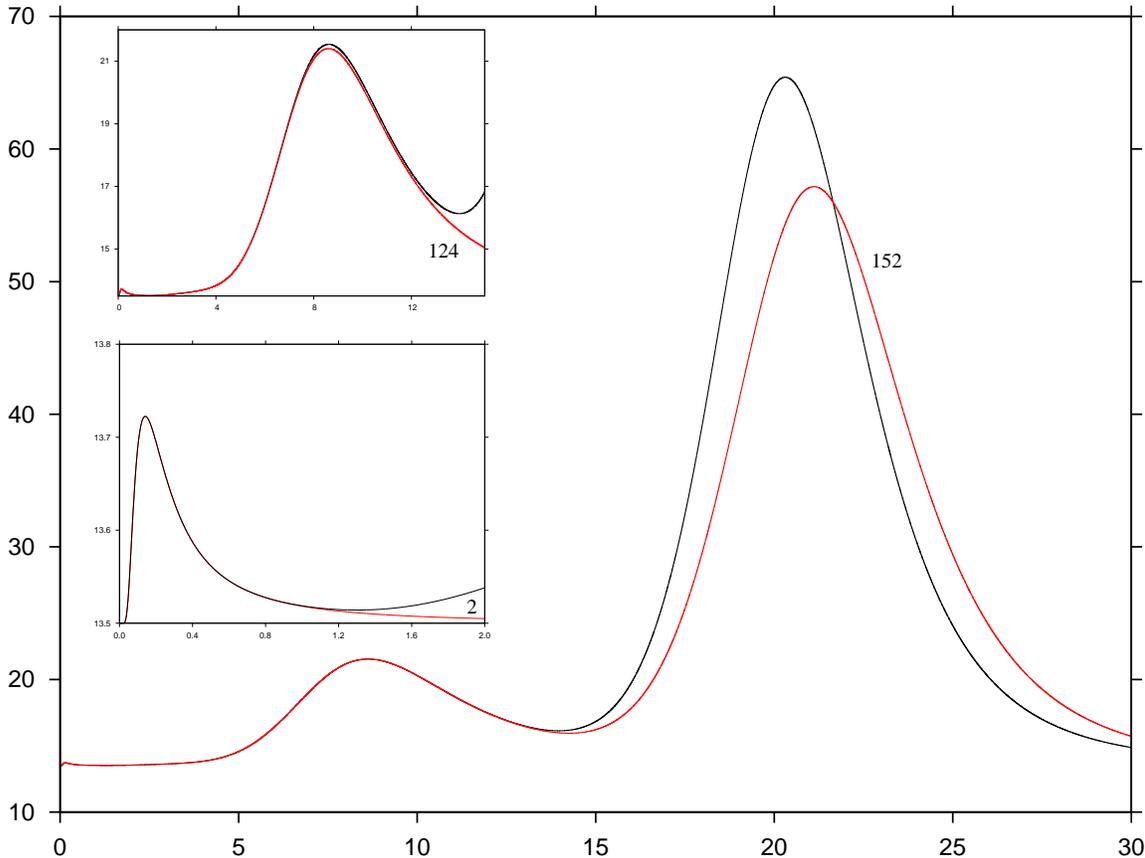}
}
\caption{Heat capacity analogue for the ML landscape defined for the dataset using only the
three initial interatomic distances with three hidden nodes.
The insets illustrate the convergence of the two low temperature peaks.
In each plot the black curve corresponds to $C_V$ calculated from the complete
database of minima. The red curves labelled `2', `124' and `152' correspond to
$C_V$ calculated from truncated sums including only the lowest 2, 124, and 152
minima, respectively.
}
\label{fig:Cv.one.3}
\end{figure}

One example is shown in Figure \ref{fig:Cv.one.3}, which is based on the ML solution landscape
for a neural network with three hidden nodes and inputs corresponding to the three interparticle
distances at the initial geometries of all the training optimisation sequences.
These results were obtained using the  pele \cite{pele} ML interface
for the neural network formulation in Eq.~(\ref{eq:yNN}), where regularisation did not
include the weights for the bias nodes \cite{BallardSDW16}.
The superposition approach provides a clear interpretation for the peaks, which we
achieve by calculating $C_V$ from partial sums over the ML training minima, in order of
increasing $E^{\rm NN}({\bf W}_{\alpha};{\bf X}_{\rm train})$.
The first peak around $k_BT\approx 0.2$ arises from competition between the lowest
two minima. 
The second peak around $k_BT\approx 9$ is reproduced when the lowest 124 minima are included, and
the largest peak around $k_BT\approx 20$ appears when we sum up to minimum 153.
The latter solution exhibits one particularly small Hessian eigenvalue, producing a
relatively large configurational entropy contribution, which increases with $T$.
The harmonic approximation will break down here, but nevertheless serves to highlight the
qualitative difference in character of minima with exceptionally small curvatures.

In molecular systems competition between alternative low energy structures often accounts for
$C_V$ peaks corresponding to  solid-solid transitions, and analogues of this situation may well
exist in the ML scenario.
Some systematic shifts in the $C_V$ analogue could result from the
density of local minima on the ML landscape (the landscape entropy \cite{SciortinoKT00,BogdanWC06,MengABM10,Wales10b}).
Understanding such effects might help to guide the construction of improved predictive tools
from combinations of fitting solutions.
Interpreting the ML analogues of molecular structure and interconversion rates between
minima might also prove to be insightful in future work.

A subsequent study investigated the quality of the predictions using two of the three 
interatomic distances, $r_{12}$ and $r_{13}$, 
and the effects of memory, in terms of input data from successive configurations chosen systematically from each geometry optimisation sequence \cite{DasW17}.
The same database of LBFGS minimisation sequences for the triatomic cluster was used here,
divided randomly into training and testing sets of equal size (5,000 sequences each).
The quality of the classification prediction into the four possible outcomes can be
quantified using the area under curve (AUC) for receiver operating characteristic (ROC) plots \cite{HastieTF09}.
ROC analysis began when radar receiver operators needed to distinguish signals corresponding to
aircraft from false readings, including flocks of birds.
The curves are plots of  the true positive rate, $T_{\rm pr}$, against the false positive rate, $F_{\rm pr}$,
as a function of the threshold probability, $P$, for making a certain classification.
Here, $P$ is the threshold at which the output probability $p^{\rm NN}_{0}({\bf W};{\bf X})$ 
is judged sufficient to predict that a minimisation would converge to the equilateral triangle, so that
\begin{eqnarray}
  T_{\rm pr}({\bf W}; {\bf X}; P) & =  & \sum_{d=1}^{N_{\rm data}} \delta_{c(d),0} \Theta(p^{\rm NN}_0({\bf W};{\bf X})-P)  \Big/ \sum_{d=1}^{N_{\rm data}}  \delta_{c(d),0}, \nonumber \\
  F_{\rm pr}({\bf W}; {\bf X}; P) & =  & \sum_{d=1}^{N_{\rm data}} (1-\delta_{c(d),0}) \Theta(p^{\rm NN}_0({\bf W};{\bf X})-P)  \Big/ \sum_{d=1}^{N_{\rm data}} (1-\delta_{c(d),0}),
\end{eqnarray}
where $\Theta$ is the Heaviside step function and $\delta$ is the Kronecker delta.
The area under the curve can then be obtained by numerical integration of
\begin{equation}
{\rm AUC}({\bf W}; {\bf X}) = \int_{0}^1  T_{\rm pr}({\bf W}; {\bf X}; P) dF_{\rm pr}({\bf W};{\bf X};  P).
\end{equation}
${\rm AUC}({\bf W}; {\bf X})$ can be interpreted as the probability that for two randomly chosen
data inputs, our predictions will discriminate between them correctly.
AUC values between 0.7 and 0.8 are usually considered `fair',
0.8 to 0.9, `good', and 0.9 to 1 `excellent'.

\begin{figure}
\psfrag{1.0}[cr][cr]{\fsz 1.0}
\psfrag{0.9}[cr][cr]{\fsz 0.9}
\psfrag{0.8}[cr][cr]{\fsz 0.8}
\psfrag{0.7}[cr][cr]{\fsz 0.7}
\psfrag{ 0}[tc][tc]{\fsz 0}
\psfrag{ 10}[tc][tc]{\fsz 10}
\psfrag{ 20}[tc][tc]{\fsz 20}
\psfrag{ 30}[tc][tc]{\fsz 30}
\psfrag{ 40}[tc][tc]{\fsz 40}
\psfrag{ 50}[tc][tc]{\fsz 50}
\psfrag{ 60}[tc][tc]{\fsz 60}
\begin{tabular}{cc}
%
      \begin{overpic}[width=0.25\textwidth,tics=10,angle=-90]{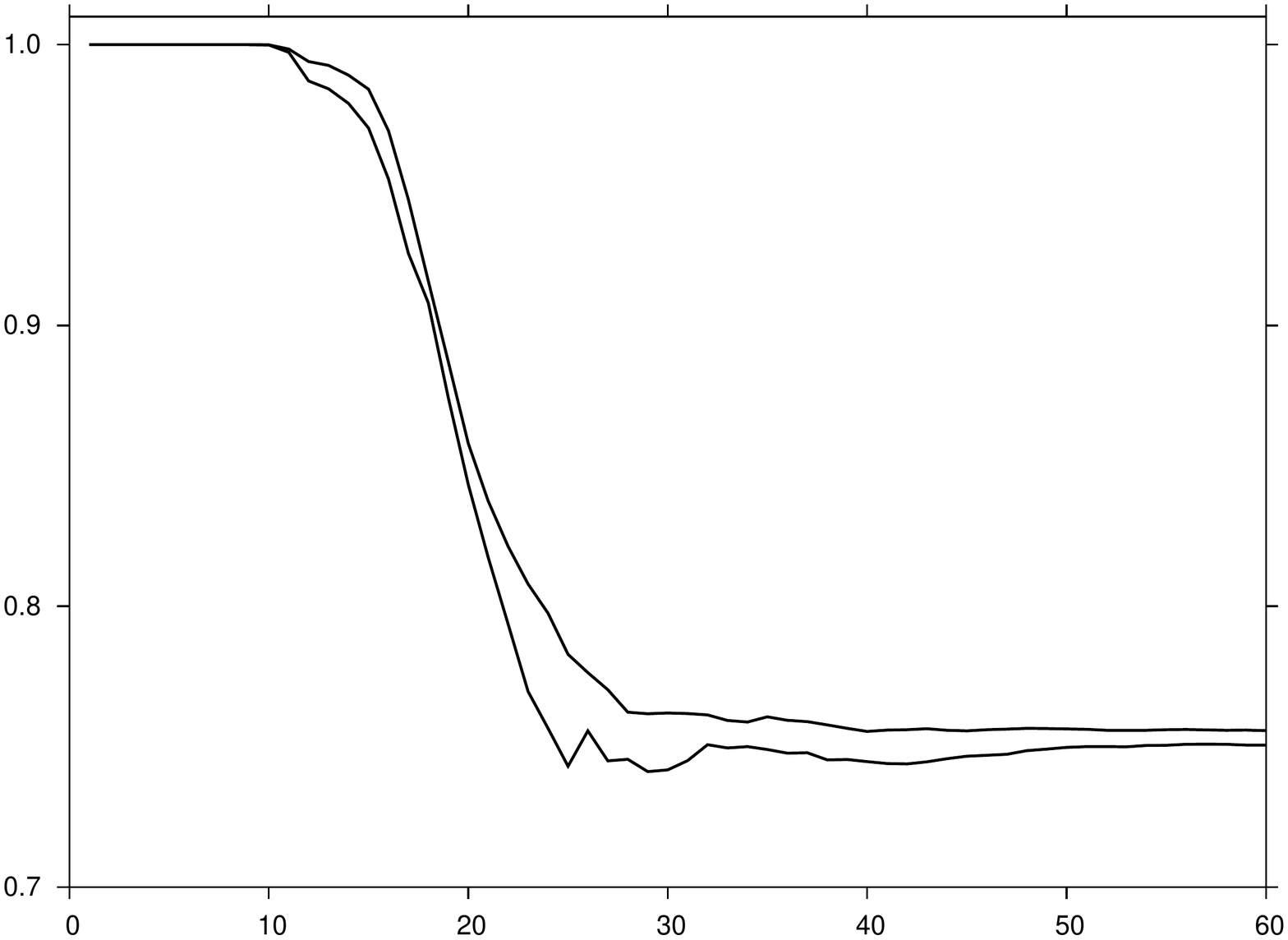}
         \put (50,55) {\fsz 3 hidden nodes} \put(-10,40){\makebox(0,0){\rotatebox{90}{AUC value}}} \end{overpic} &
      \begin{overpic}[width=0.25\textwidth,tics=10,angle=-90]{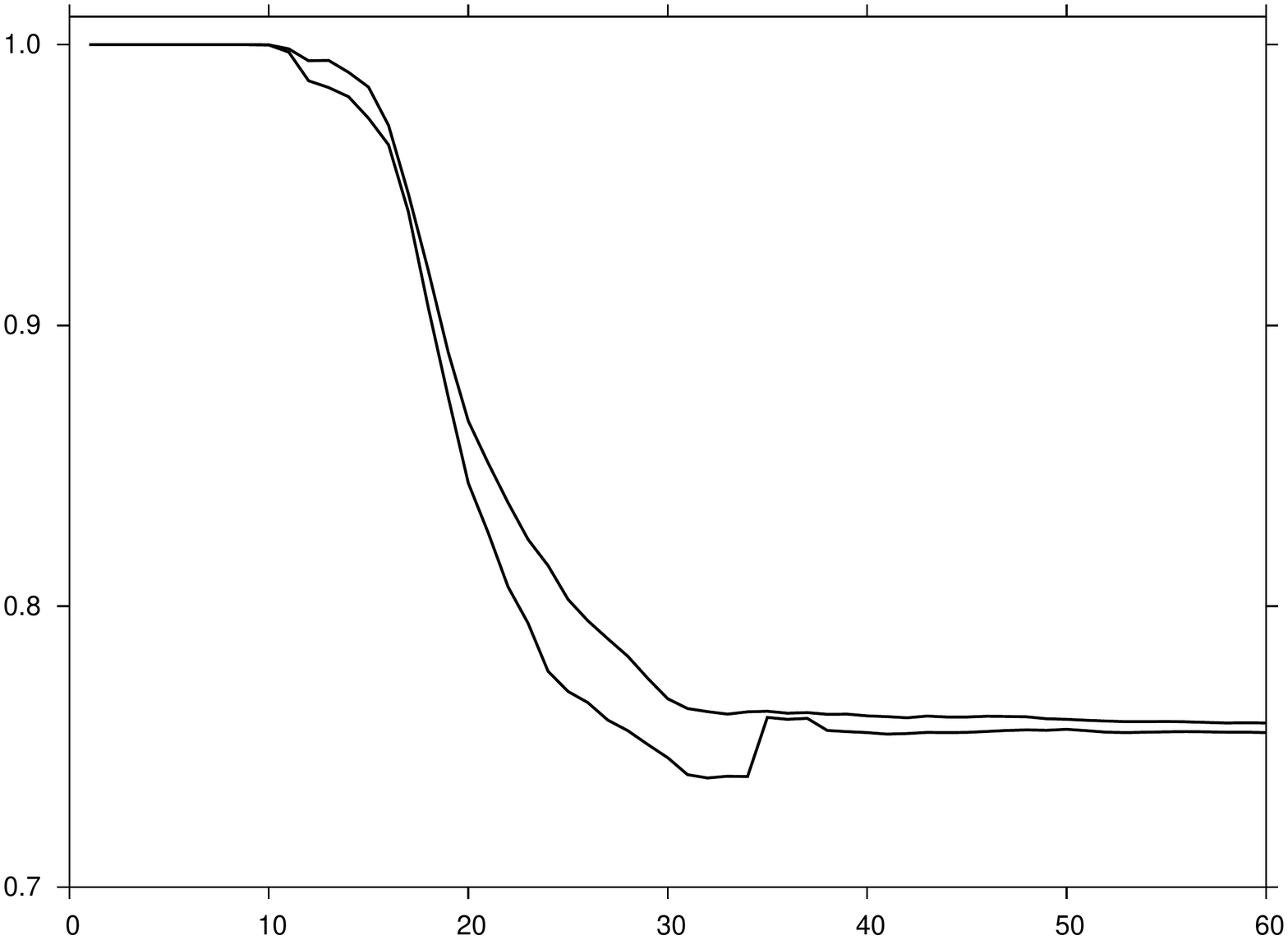}
         \put (50,55) {\fsz 4 hidden nodes} \end{overpic} \\
      \begin{overpic}[width=0.25\textwidth,tics=10,angle=-90]{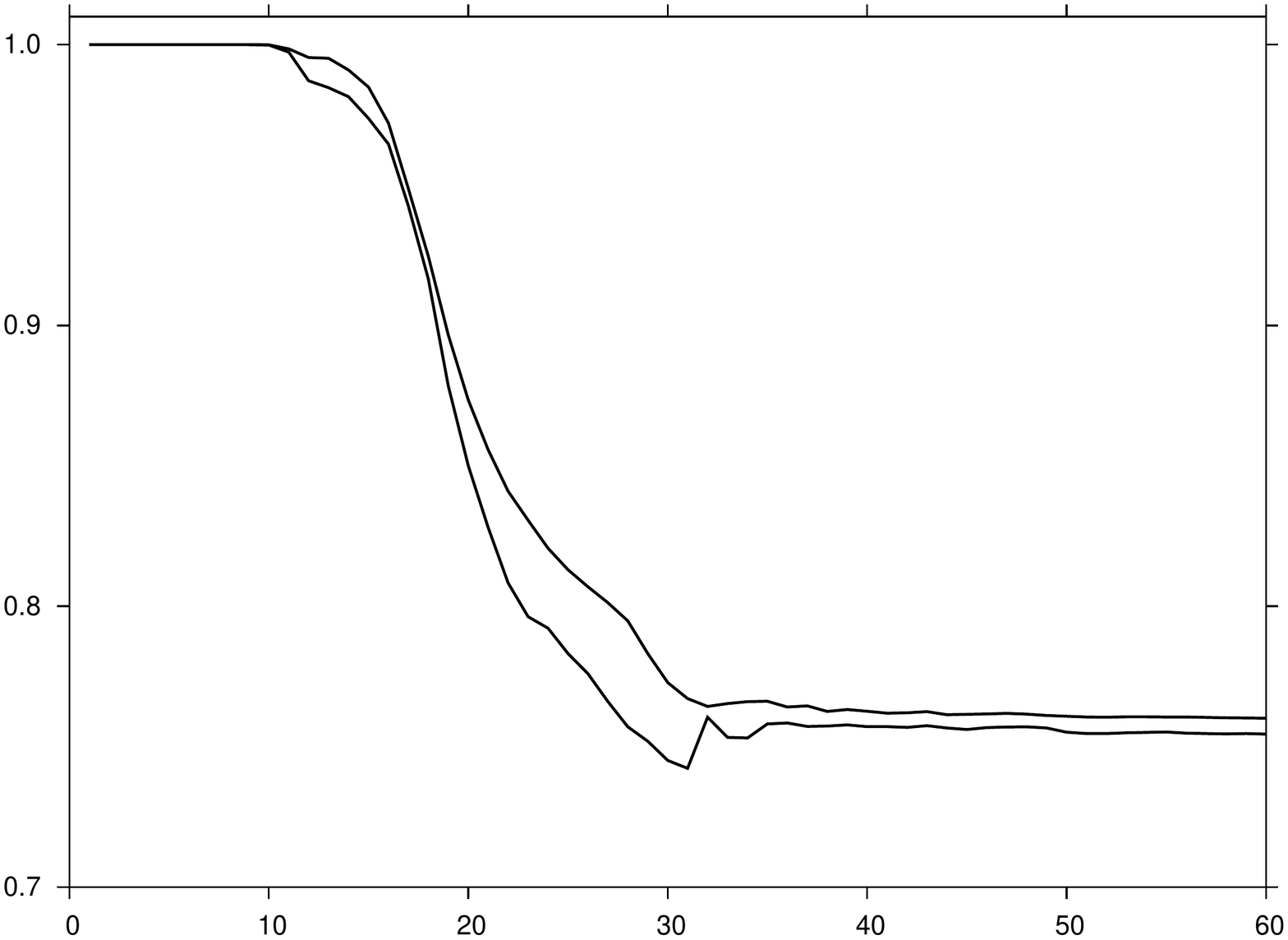}
         \put (50,55) {\fsz 5 hidden nodes} \put(60,-10){$s$ (steps from convergence)} \end{overpic} &
      \begin{overpic}[width=0.25\textwidth,tics=10,angle=-90]{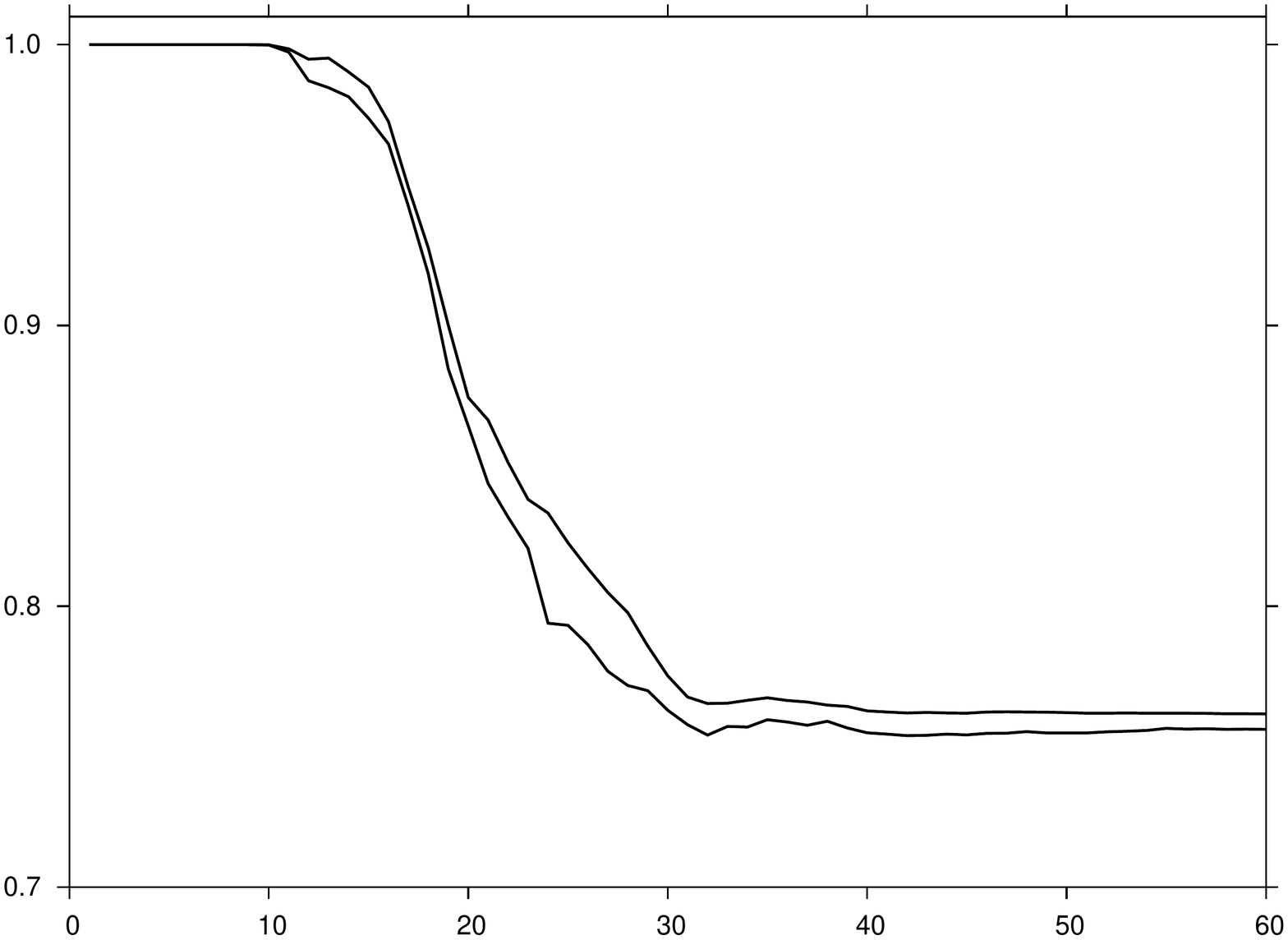}
         \put (50,55) {\fsz 6 hidden nodes} \end{overpic} \\
\end{tabular}
\bigskip
\caption{\label{fig:AUCcheck}
AUC values for 5,000 minimisation sequences in the LBFGS testing set,
evaluated using the parameters obtained for the global minimum neural network
fit with 5,000 training sequences and $\lambda=10^{-4}$.
The four panels correspond to 3, 4, 5 or 6 hidden nodes, as marked,
and the lower curve corresponds to the global minimum of 
$E^{\rm NN}({\bf W}_{\alpha};{\bf X}_{\rm train})$ with
${\bf X}_{\rm train}$ containing
$r_{12}$ and $r_{13}$ at a single configuration in each minimisation sequence,
located $s$ steps from convergence.
Each panel has a second plot of the
highest AUC value for the test data attained with any local minimum obtained in training having the
same number of inputs and hidden nodes, including results for all
the $\lambda$ values considered and for all values of $s$, from 1 to 80.
The AUC value for the global minimum with $\lambda=10^{-4}$ and the
configuration in question is included in this set, but can be exceeded by one of the
many local minima obtained over the full range of $\lambda$ and $s$.
Beyond $s$ around 60 the plots are essentially flat.
}
\end{figure}

Figure \ref{fig:AUCcheck} shows the AUC values obtained with the LBFGS database when
the input data consists of $r_{12}$ and $r_{13}$ values at different points in the 
minimisation sequence. Here the horizontal axis corresponds to $s$, the number of
steps from convergence, and the AUC values therefore tend to unity when $s$ is small,
where the configurations are close to the final minimum. Each panel in the Figure 
includes two plots, which generally coincide quite closely.
The plot with the lower AUC value is the one obtained with the global minimum 
located for $E^{\rm NN}({\bf W}_{\alpha};{\bf X}_{\rm train})$ for configurations
$s$ steps for convergence.
The AUC value in the other plot is always greater than or equal to the value for
the global minimum, since it is the maximum AUC calculated for all the local minima
characterised with the same neural net architecture and any $s$ value.
Minimisation sequences that converge in fewer than $s$ steps are padded with the initial configuration
for larger $s$ values, which is intended to present a worst case scenario.

Figure \ref{fig:AUCcheck} shows that the prediction quality decays in a characteristic fashion
as configurations move further from the convergence limit.
It also illustrates an important result, namely that the performance of the global minimum
obtained with the training data is never surpassed significantly by any of the other
local minima, when all these fits are applied to the test data.
Hence the global minimum is clearly a good place to start if we wish to make predictions,
or perhaps construct classification schemes based on more than one local minimum of the
ML landscape obtained in training.
The corresponding fits and AUC calculations were all rerun to produce 
Figure \ref{fig:AUCcheck} using the fitting function defined in Eq.~\ref{eq:yNN}
and regularisation over all variables,
for comparison with the simplified bias weighting employed in \cite{DasW17}.
There is no significant difference between our results for the two schemes.

\section{Non-Linear Regression}
\label{sec:regression}

Regression is perhaps the most well-known task in machine learning,
referring to any process for estimating the relationships between dependent and independent variables.
As we show
in this section, even a relatively simple non-nonlinear regression problem
leads to a rich ML landscape. As in the standard regression scenario, we
consider a set of $N_{\rm data}$ data points $D = ((x_1,t_1),\cdots,(x_{N_{\rm data}},t_{N_{\rm data}}))$, and a
model $y(x; \vq)$ that we wish to fit to $D$ by adjusting $M$ parameters $\vq =
(q_1, q_2, \cdots, q_M)$. In this example, we investigate the following
non-linear model:
 \begin{equation}
  y(x; \vq) = e^{-q_1 x} \sin(q_2 x + q_3) \sin(q_4 x + q_5).
  \label{eq:model}
 \end{equation}
Our regression problem is one-dimensional ($x$ is a scalar), with a five-dimensional vector $\vq$ that parameterises the model.

We performed regression on the above problem, with a dataset $D$ consisting of
$N_{\rm data}=100$ data points with $x_i$ values sampled uniformly in $[0, 3\pi ]$, and
corresponding $t_i$ values given by our model with added Gaussian white noise
(mean zero, $\sigma=0.02$):
\begin{equation}
t_i = y(x_i; \vq^\star) + \text{noise},
\end{equation} 
with a particular ad hoc parameter choice $\vq^\star  = (0.1, 2.13, 0.0, 1.34, 0.0)$. The cost function we minimise is a standard sum of least squares:
\begin{equation}
E({\bf \vq}) = \sum_{i=1}^{N_{\rm data}} \left[t_i - y(x_i; \vq)\right]^2.
\label{eq:least_squares}
\end{equation}
The objective of this regression problem is to find a best-fit mapping from input (${x}$) to target variables (${t}$).  
Intuitively, we expect minimisation of Eq.~(\ref{eq:least_squares})
with respect to the parameters ${\bf q}$ to yield an
optimal value $\vq \approx \vq^\star$. However, since $E$ is a non-convex
function of $\vq$, there are multiple solutions to the equation $\nabla E({\bf \vq}) =
{\mathbf 0}$ and hence the outcome depends on the minimisation
procedure and the initial conditions. 

\begin{figure}[!ht]
 \begin{centering}
 \includegraphics[width=0.6\textwidth]{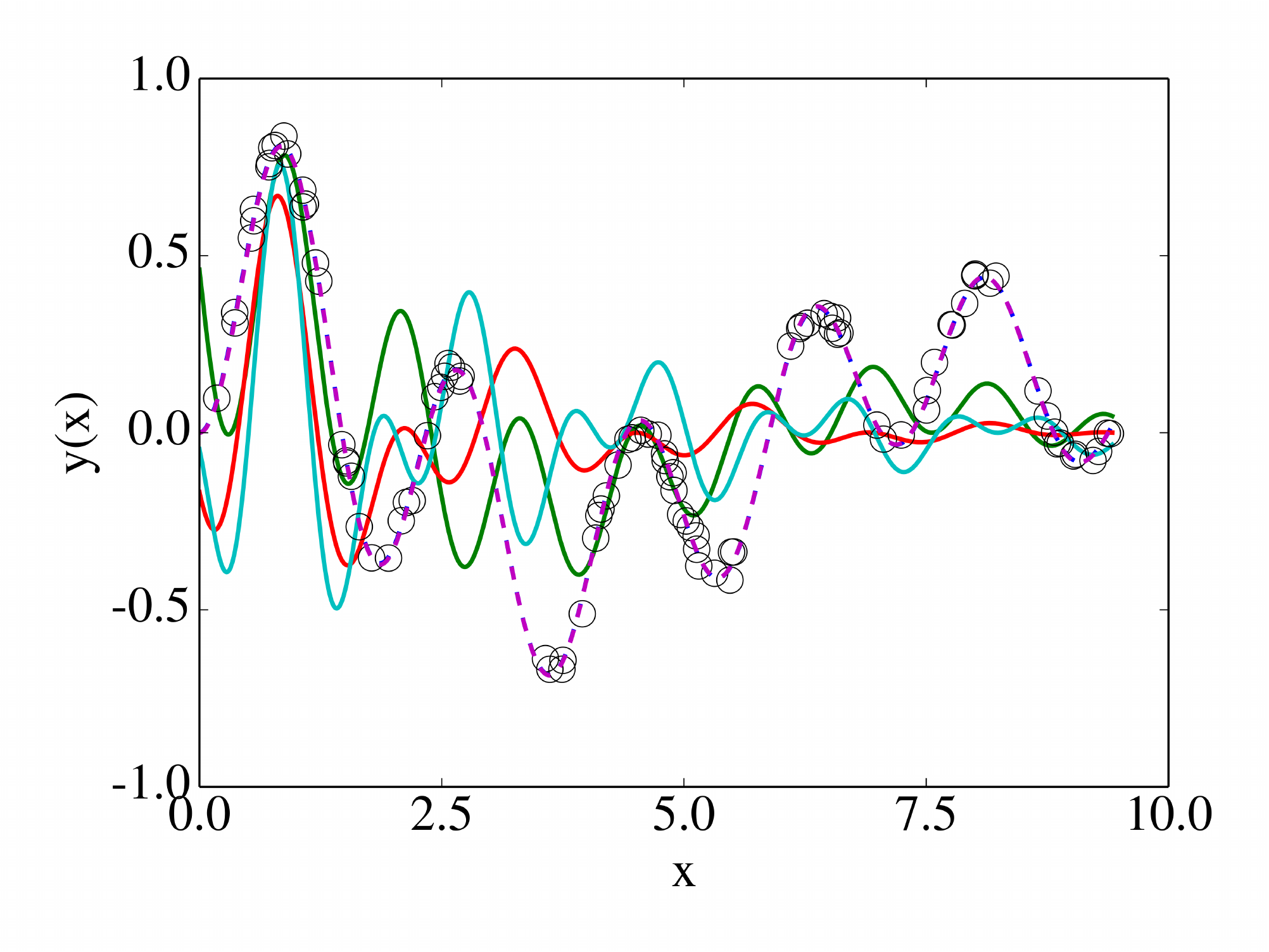}
  \caption{Results from nonlinear regression for the model given by
Eq.~(\ref{eq:model}): the global minimum of the cost function (dashed line) is
plotted with various local minima solutions (solid lines) and the data used for
fitting (black circles). The model used to generate the data is
indistinguishable from the curve corresponding to the global minimum. }
  \label{fig:curve}
  \end{centering}
  \end{figure}

\begin{figure}[!ht]
 \includegraphics[width=0.5\textwidth]{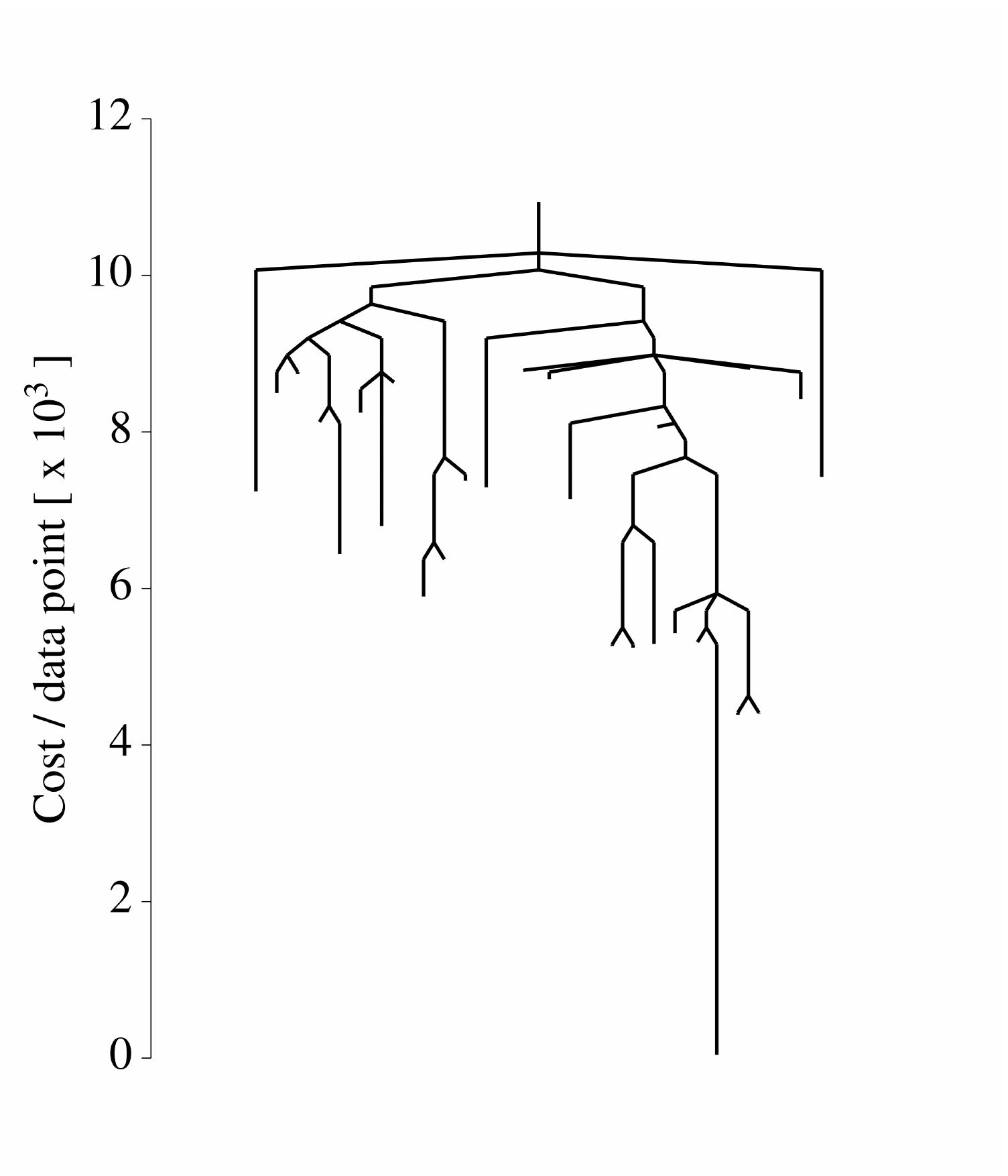}
  \caption{Disconnectivity graph for a nonlinear regression cost function [Eq.~(\ref{eq:model})].  
  Each branch terminates at a local minimum at the value of $E({\bf \vq})$ for that minimum.  
  By following the lines from one minimum to another, one can read off the energy barrier on the minimum energy path connecting them.
  }
  \label{fig:dg1}
  \end{figure}

We explored the landscape described by $E(\vq)$, and display the various
solutions in Fig.~\ref{fig:curve} alongside our data and $y(x;\vq=\vq^\star)$. In this case, the global minimum is a fairly accurate
representation of the solution used to generate $D$. 
However, $88$ other solutions were found which do not accurately represent the
data, despite being valid local minima. In Fig.~\ref{fig:dg1} we show the
disconnectivity graph for $E({\vq})$.  
Here the vertical axis corresponds to $E({\vq})$, and branches terminate at the values for
corresponding local minima, as in \S \ref{sec:LJAT3}.
The graph shows that the energy of the global minimum
solution is separated from the others by an order of magnitude, and is clearly
the best fit to our data (dotted curve in Fig.~\ref{fig:curve}). The barriers
in this representation can be interpreted in terms of transformations between
different training solutions in parameter space, and could be indicative of the
distinctiveness of the minima they connect. The minima of this landscape were found 
by the basin-hopping method\cite{lis87,lis88,WalesD97}, as described in \S \ref{sec:energy-landscapes}.

\section{Digit Recognition using a Neural Network}
\label{sec:NN}


The next machine learning landscape we explore here is another artificial
neural network, this time trained for digit recognition on the MNIST
dataset \cite{lecun1998}. Our network architecture consists of $28\times28$
input nodes corresponding to input image pixels, a single hidden layer with
$10$ nodes, and a softmax output layer of $10$ nodes, which represent the
10 digit classes. This model contains roughly $8000$ adjustable parameters, which quantify the weight of given nodes in
activating one of their successors. The cost function we optimise for this
classification example is the
same multinomial logistic regression function that is described above in
\S \ref{sec:LJAT3}, which is standard for classification problems. 
Here `logistic' means that the dependent variable (outcome) is a category,
in this case the assignment of the image for a digit,
and `multinomial' means that there are more than two possible outcomes.
An $L^2$ regularisation term was again added to the cost function, as
described in \S \ref{sec:LJAT3}.
Unless otherwise mentioned, all the
results described below are for a regularisation coefficient of $\lambda = 0.1$. 

\begin{figure}[!ht]
 \includegraphics[width=0.6\textwidth]{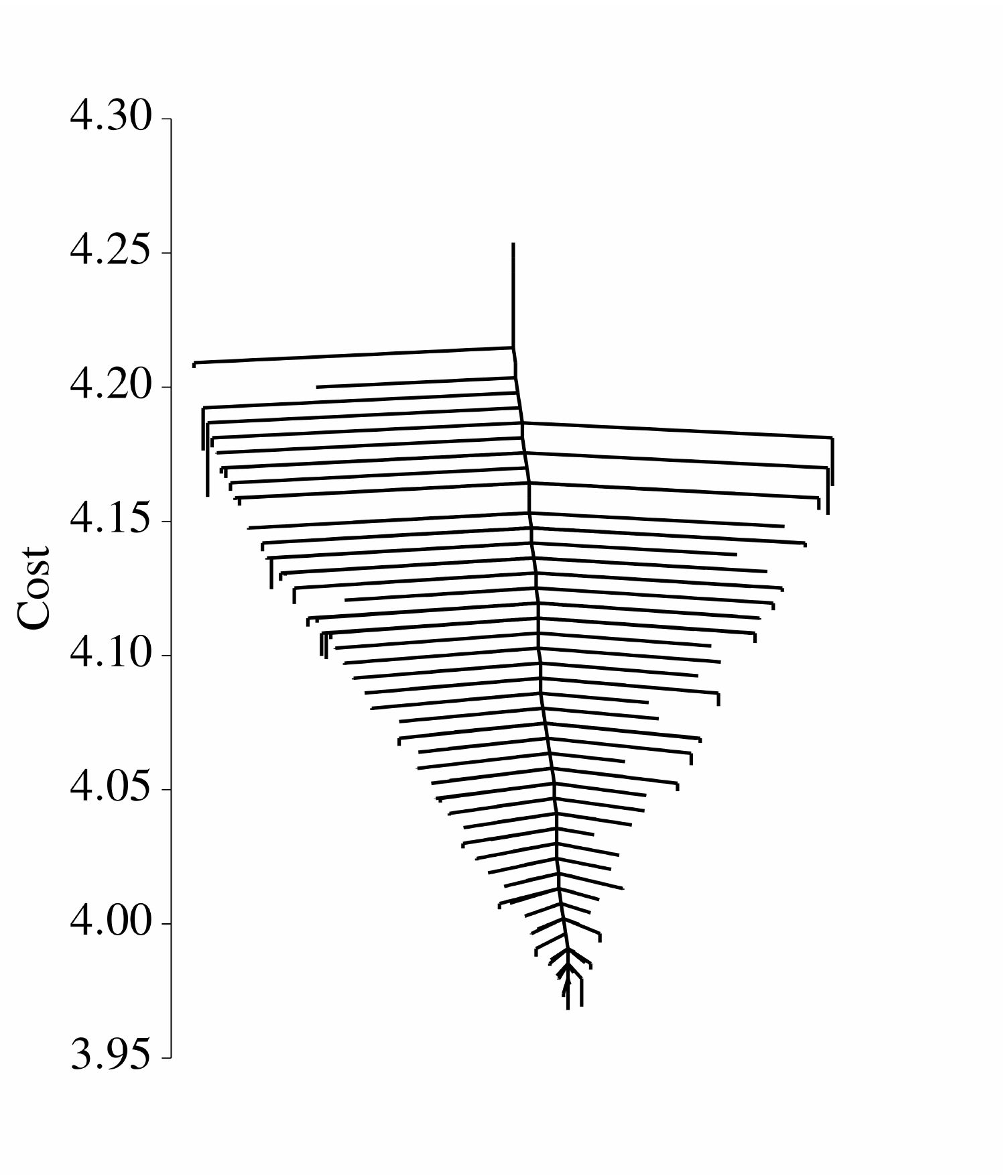}
  \caption{Disconnectivity graph of neural network ML solutions for digit recognition.}
  \label{fig:dg2}
  \end{figure}

\begin{figure}[!ht]
   \includegraphics[width=0.5\textwidth]{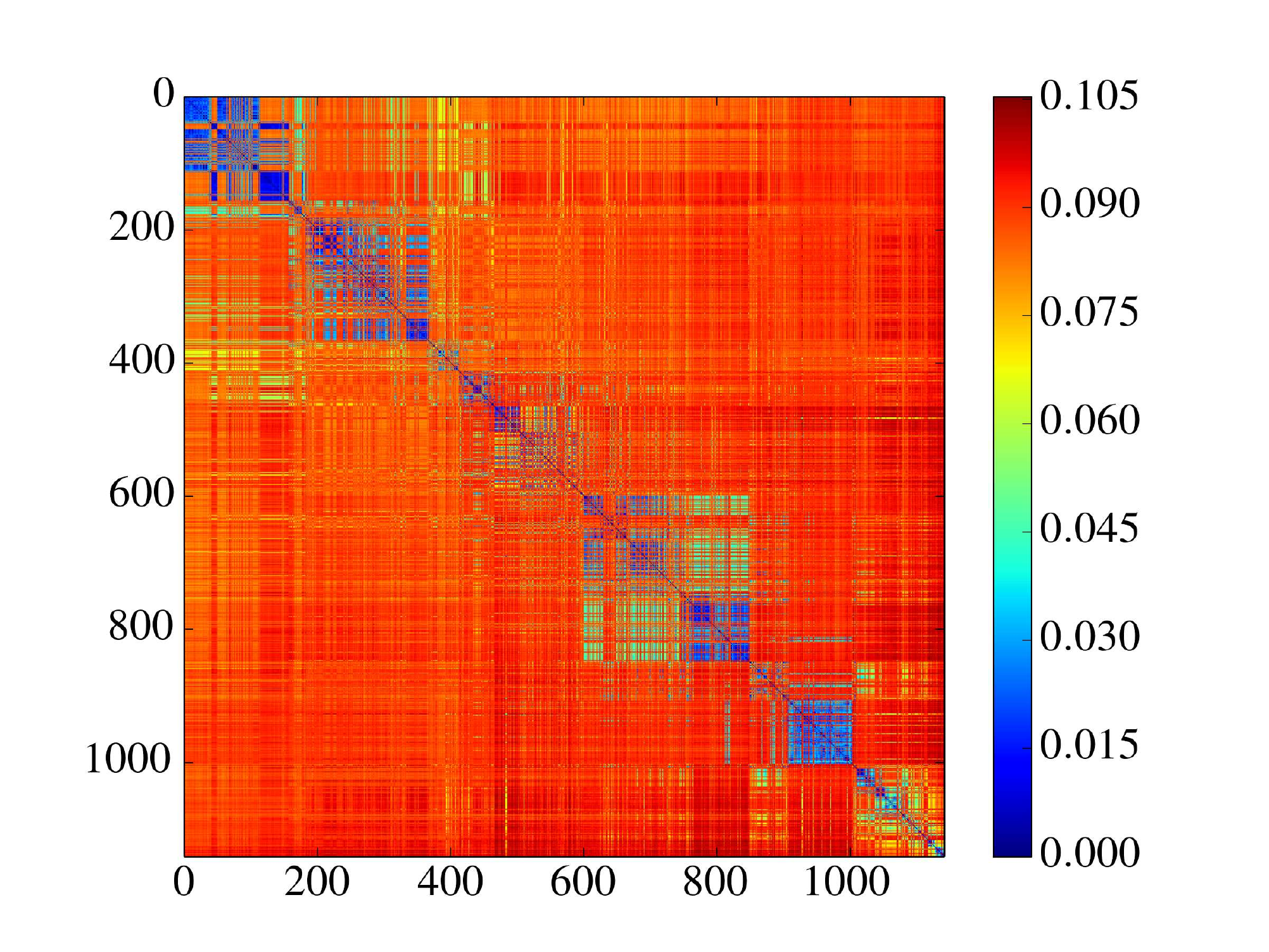}
  \caption{Misclassification heat map for various solutions to the digit 
recognition problem on the ML landscape. This map displays
the degree of similarity of any pair of minima based upon correct test set
classification. See text for details.}
  \label{fig:hamming}
  \end{figure}

\begin{figure}[!ht]
 \includegraphics[width=0.5\textwidth]{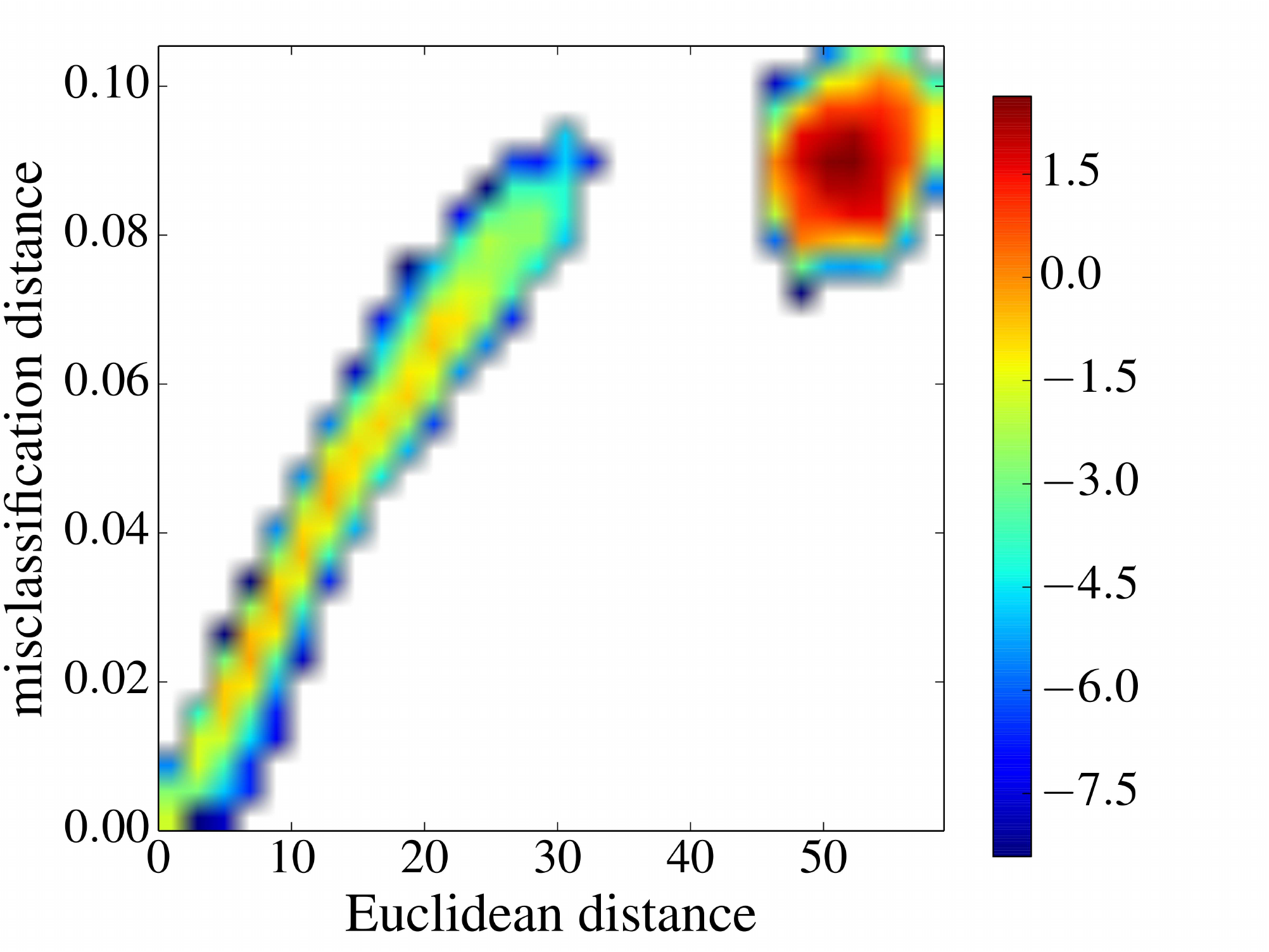}
  \caption{Joint probability density plot of minimum-minimum Hamming distance
and geometric distance in parameter space. The distance metric in
parameter space is correlated with the misclassification distance between
minima. The probability density is coloured on a log scale.}
  \label{fig:dist_vs_ham}
  \end{figure}

The neural network defined above is quite small, and is not intended to compete with
well-established models trained on MNIST~\footnote{See the MNIST database:
http://yann.lecun.com/exdb/mnist}. Rather, our goal in this Perspective is 
to gain insight into the landscape. This aim is greatly
assisted by using a model that is smaller, yet still behaves similarly to more
sophisticated implementations. To converge the disconnectivity graph, Fig.~\ref{fig:dg2},
in particular the transition states, the model was trained on $N_{\rm data}=1000$ data
points.  The results assessing performance, Figs.~\ref{fig:hamming}
and \ref{fig:dist_vs_ham}, were tested on $N_{\rm data}=10,000$ images. 

We explored the landscape of this network for  several different values of the
$L^2$ regularisation parameter $\lambda$. The graph shown in
Fig.~\ref{fig:dg2}, with $\lambda=0.01$, is representative of all the others:
we observe a single funnel landscape, where, in contrast to the
nonlinear regression example, all of the minima occur in a narrow range of
energy (cost) values.  This observation is consistent with recent work suggesting
that the energies of local minima for neural networks are spaced exponentially
close to the global minimum \cite{Dauphin2014, Bray2007} with 
the number of variables (number of optimised parameters or dimensionality) of the system. 

We next assess the performance profile of the NN minima by calculating the
misclassification rate on an independent test set. Judging by average
statistics the minima seem to perform very similarly: the fraction $f$ of
misclassified test set images is comparable for most of them, with $0.133 \le f
\le 0.162$ (mean $\bar f = 0.148$, standard deviation $\sigma_f = 0.0043$). This observation is also
consistent with previous results where local minima were found to perform quite
similarly in terms of test set error \cite{LeCun2015,sagun2014explorations}. When looking beyond
average statistics, however, we uncover more interesting behaviour. To this end
we introduce a {\it misclassification distance} $\ell$ between pairs of minima,
which we define as the fraction of test set images that are misclassified by
one minimum but not both~\footnote{The misclassification distance can also be
viewed as the Hamming distance between misclassification vectors of the two
minima in question.}. A value $\ell_{ij}=0$ implies that all images are
classified in the same way by the two minima; a value
$\ell_{ij}=\ell^{max}_{ij} = f_i + f_j$ implies that every misclassified image
by $i$ was correctly classified by $j$, and every misclassified image by $j$
was correctly classified by $i$.
In Fig.~\ref{fig:hamming} we display the matrix $\ell_{ij}$, which shows that the
minima cluster into groups that are self-similar, and distinct from other
groups. So, although all minima perform almost identically when considering the
misclassification rate alone, their performance looks quite distinct when
considering the actual sets of misclassified images. We hypothesise that this
behaviour is due to a saturated information capacity for our model. This small
neural network can only encode a certain amount of information during the
training process. Since there are many training images, there is much more
information to encode than it is possible to store. The clustering of minima
in Fig.~\ref{fig:hamming} then probably reflects the differing information content that each
solution retains. Here it is important to remember that each of the minima were
trained on the same set of images; the distinct minima arise solely from the
different starting configurations prior to optimisation. 

The misclassification similarity can be understood from the underlying ML landscape. 
We investigated correlations between misclassification distance and distance in parameter space. 
In Fig.~\ref{fig:dist_vs_ham} we display the joint distribution of the misclassification distance and 
Euclidean distance ($L^2$ norm) between the parameter values for each pair of minima. 
We see that for a range of values these two measures are highly correlated, indicating that the misclassification 
distance between minima is determined by their proximity on the underlying landscape. 
Interestingly, for very large values of geometric distance there is a large (yet seemingly random) 
misclassification distance. 
 
The seemingly random behaviour could possibly be the result of 
symmetry with respect to permutation of neural network parameters. 
There exist a large number of symmetry operations for the parameter space 
that leave the NN prediction unchanged, 
yet would certainly change the $L^2$ distance with respect to another reference point. 
A more rigorous definition of distance (currently unexplored), would take such symmetries into account.
There are at least two such symmetry operations~\cite{Bishop06}. 
The first of these results from the antisymmetry of the $\tanh$ activation function: 
inverting the sign of all weights and bias leading into a node will 
lead to an inverted output from that node. 
The second symmetry is due to the arbitrary labelling of nodes: swapping the 
labels of nodes $i$ and $j$ within a given 
hidden layer will leave the output of a NN unchanged.

\section{Network Analysis for Machine Learning Landscapes}
\label{sec:network_of_minima}

The landscape expressed in terms of a connected database of local minima and
transition states can be analysed in terms of network properties
\cite{doye2002network,doye2005characterizing,MehtaCCKW16}. 
The starting point of such a description is the observation 
that for a potential energy function with continuous degrees of freedom, 
each minimum is connected to other minima through steepest-descent paths mediated by transition states.  
Hence, one can construct a network \cite{Networks}
in a natural way, where each minimum corresponds to a node.
If two minima are connected via a transition state then there is an edge between them.
In this preliminary analysis we consider an unweighted and undirected network; this approach will be extended 
in the future to edge weights and directions that are relevant to kinetics. 
In this initial analysis we are only interested in whether or not two minima are connected,
and multiple connections make no difference. 
The objective of working with unweighted and undirected networks is to first focus on the global structure of the landscape,
providing the foundations for analysis of how emergent thermodynamic and kinetic properties are encoded
in future work.

After constructing the network we can analyse properties such as average shortest path length,
diameter, clustering coefficients, node degrees and their distribution. For an unweighted and undirected 
network, the shortest path
between a pair of nodes is the path that passes through the fewest edges. 
The number of edges on the shortest path is then the shortest path length between the pair 
of nodes, and the average shortest path length is the average
of the shortest path lengths over all the pairs of nodes. The diameter of a network is the path length of the longest 
shortest path. For the network of minima,
the diameter of the network is the distance, in terms of edges, between the pair of minima that are farthest apart.
The node degree is the number of directly connected neighbours.

\begin{figure}[!ht]
 \includegraphics[width=0.5\textwidth]{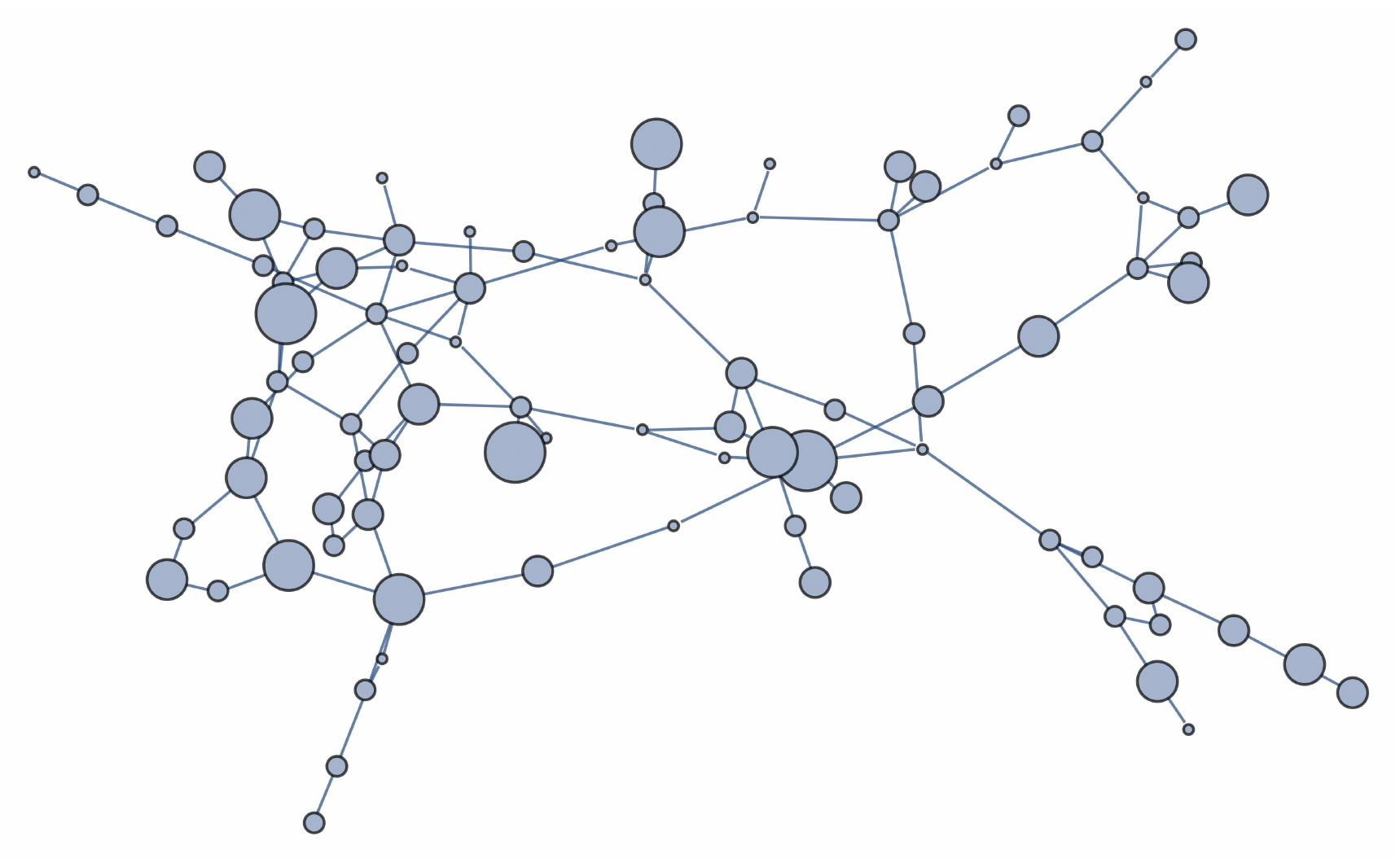}
  \caption{Network of minima for the nonlinear regression cost function [Eq.~(\ref{eq:model})]. Each node corresponds
  to a minimum. There is an edge between two minima if they are connected by at least one transition state. The size of the nodes
  is proportional to the number of neighbours.
  }
  \label{fig:nonlinear-reg-network}
  \end{figure}

For the non-linear regression model, we found $89$ minima and $121$ transition states. The resulting network 
consists of $89$ nodes and $113$ edges (Figure \ref{fig:nonlinear-reg-network}). Although this is a small 
network we use it to introduce the analysis.
The average node degree is $2.54$ and the average shortest path length 
is $6.0641$.  The network diameter, i.e.~the longest shortest path, 
is $15$. Hence, on average a minimum is around $6$ 
steps away from any other minimum, and the pair farthest apart are separated by 15 steps.
Thus, a minimum found by a naive 
numerical minimisation procedure may be on average 6 steps, and in the worst case 15 steps, 
from the global minimum. Both the average shortest path and network diameter of this network 
are significantly larger than for a random network of an equivalent size \cite{Strogatz01}.

\begin{figure}[!ht]
 \includegraphics[width=0.5\textwidth]{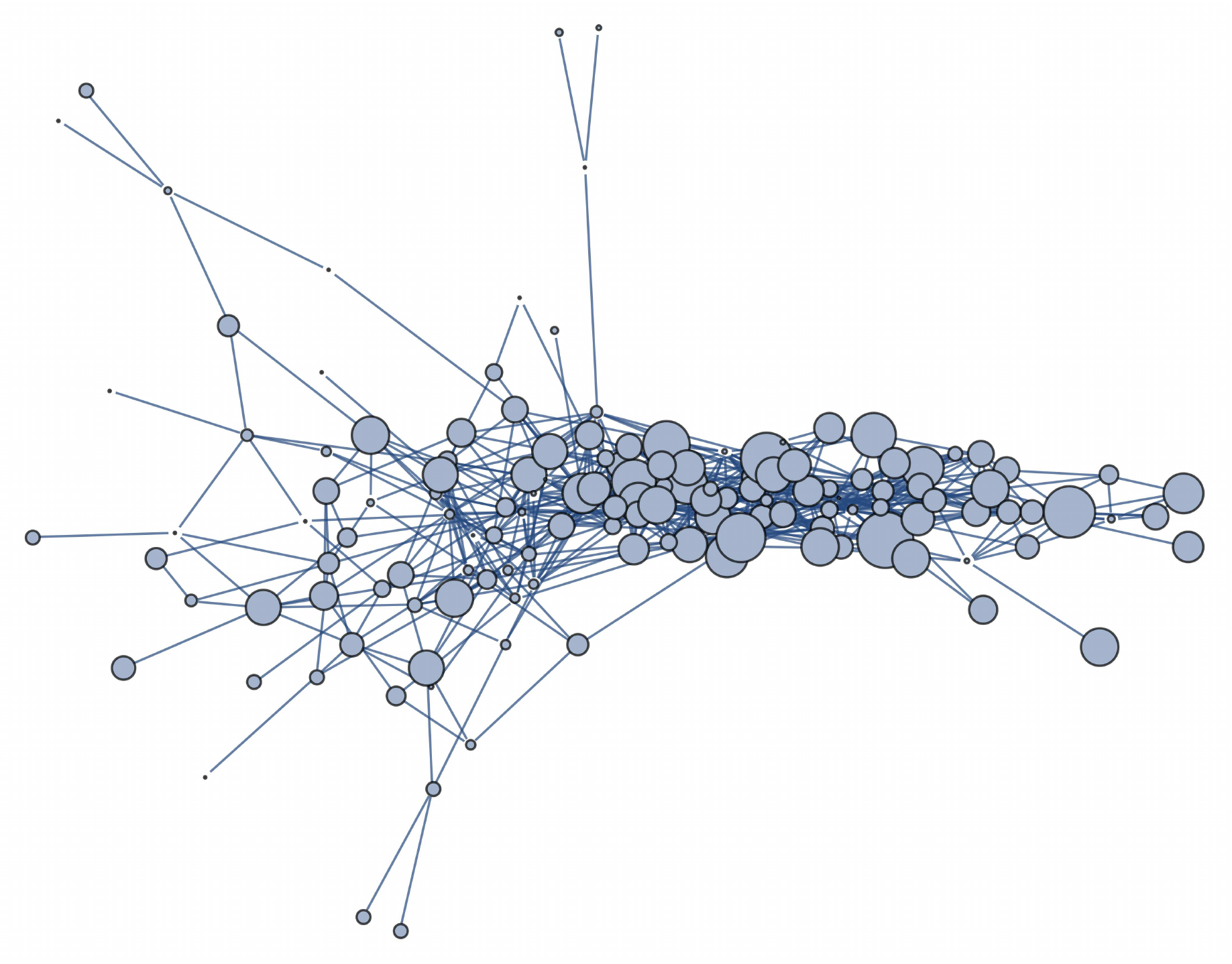}
  \caption{Network of minima for the NN model cost function [Eq.~(\ref{eq:model})]
applied to digit recognition. The size of the nodes is proportional to the number of neighbours.
  }
  \label{fig:nn-network-of-minima}
  \end{figure}

The network of minima for the neural network model in Sec.~\ref{sec:NN} has $142$ 
nodes and $643$ edges (Figure \ref{fig:nn-network-of-minima}).
The nodes have on average $9.06$ nearest neighbours,
the average shortest path is $3.18$, and the network diameter is $8$.
Hence, a randomly selected minimum is on an average only $3$ steps away from any other minimum
in terms of minimum-transition state-minimum triples. Therefore, on average, the global minimum is only 3 steps away from 
any other local minimum. The networks of minima defined by molecules such as Lennard-Jones clusters, 
Morse clusters, and the Thomson problem 
have also been shown to have small (of order $O(\log[\mbox{number of minima}])$ average shortest path lengths, meaning that
any randomly picked local minimum is only a few steps away from the global minimum 
\cite{doye2002network,doye2005characterizing,MehtaCCKW16, MorganEtal2016}.
Moreover, these networks exhibit small-world behaviour \cite{watts1998collective}, i.e.~the 
average shortest path lengths of these networks are small and similar to equivalent size random networks,
whereas their clustering coefficients are significantly larger than those of equivalent size random networks.
We have conjectured that the
small-world properties of networks of minima are closely related to the single-funnel nature of 
the corresponding landscapes \cite{walesdmmw00,Doye02,CarrW08}.
Some networks also exhibit scale-free properties \cite{barabasi1999emergence}, where the node-degrees 
follow a power-law distribution. In such networks only a few nodes termed hubs have most of the 
connections, while most other nodes have only a small number. 
%
%

The benefits of analysing network properties of machine learning landscapes may be two-fold.
One can attempt to construct more efficient and tailor-made algorithms to find the global minimum of machine learning problems 
by exploiting these network properties, for example, by navigating the `shortest path' to the global minimum.
We also obtain a quantitative description of the distance between a typical minimum from the best fit solution.

In the future, we plan to study further properties of the networks of minima 
for a variety of artificial neural networks and test the small-world
and scale-free  properties.
We hope that these results may be useful in constructing 
algorithms to find the global minimum of non-convex machine learning cost functions.
 
\section{The p-Spin Model and Machine Learning}
\label{sec::pspin}

Many machine learning problems are solved via some kind of optimisation that
makes use of gradient based methods, such as the stochastic gradient descent.
Such algorithms utilise the local geometry of the landscape, and eventually
iterations stop progressing when the norm of the (stochastic) gradients approach
zero. This leads to the following general question: for a real valued
function, what values do the critical points have? The answer certainly
depends on the structure of the function we have at hand. In this section, we
will examine two classes of functions: the Hamiltonian of the $p$-spin
spherical glass, and the loss function that arises in the optimisation of a
simple deep learning problem. 
In this section, the deep learning problem is the same as the one introduced in Section \ref{sec:NN} with a larger hidden layer and 
zero regularisation.
The functions have very different structures, and at
first sight they do not appear to resemble one another in any meaningful way. 
However, we find that the two systems may exhibit similar characteristics in terms
of the values of their critical values, in spite of the apparent differences.

\subsection{Concentration of critical points of the p-Spin Model}

The Hamiltonian of a mean-field spin glass is a homogeneous polynomial of a
given degree, where the coefficients of the polynomial describe the nature and
strength of the interaction between the spins. Since the polynomial is
homogeneous, a common choice is to restrict the variables (spins) to the unit
sphere. We investigate what can be said about the minimisation problem if we
choose the coefficients of this polynomial at random and independently from the
standard normal distribution.

First, we define the notation for the rest of the section. We consider real
valued polynomials, $H(\vect{w})$, where $\vect{w}$ is the vector of variables
of $H$; the degree of the polynomial $H$ is $p$. We define the dimension
of the polynomial by the length of the vector $\vect{w}$, so if
$\vect{w}=(w_1,...,w_N)$ then $H$ is an $N$- dimensional
polynomial. A degree $p$ polynomial is homogeneous polynomial if
it satisfies the following condition for any real $t$:
\begin{equation}
H(t\vect{w})=H((tw_1, ..., tw_N))=t^pH(\vect{w})
\end{equation}
Finally, a degree $p$ polynomial of $N$ variables will have $N^p$
coefficients (some of which may be zero). The coefficients will be
denoted $x_{i_1,...,i_p}$, where each index runs from $1$ to $N$.

Having defined the notation, we now clarify the connection between polynomials
and spin glasses. Suppose the vector, $\vect{w}=(w_1,...,w_N)$, describes the
states of $N$ Ising spins that are $+1$ or $-1$. Then $\sum w_i^2 = N$, so that
the distance to the origin is $\sqrt{N}$. The continuous analogue of this model
is a hypercube embedded in a sphere of radius $\sqrt{N}$. Therefore, we can
interpret the Hamiltonian of a spherical, $p$-body, spin glass by a homogeneous
polynomial of degree $p$. This formulation for spin systems has been studied
extensively in \cite{auffinger2013random, auffinger2013complexity,
auffinger2017energy}. From here on, we will explicitly denote the dimension and
the degree of the polynomial in a subscript.

The simplest case is when the degree is $p=1$ and the polynomial (Hamiltonian) becomes
\begin{equation}
H_{N, 1}(\vect{w})=\sum_{i=1}^N w_ix_{i},
\end{equation}
where the spins $(w_1, \dots, w_n) \equiv \vect{w} \in \mathbb{R}^N $ are
constrained to the sphere \textit{i.e.} $\sum_{i=1}^{N} w_i^2 = N$, where
$N$ is the number of spins. The coefficients $x_i\sim \mathcal{N}(0,1)$ are
independent and identically distributed 
standard normal random
variables. For $p=1$ there exist only two stationary points, one minimum and one maximum.

When $p=2$ the polynomial becomes
\begin{equation}
H_{N, 2}(\vect{w})=\sum_{i,j=1}^Nw_iw_jx_{ij}.
\end{equation}
This is a simple quadratic form with $2N$ stationary points located at the
eigenvectors of the matrix $\bf X$ with elements $X_{ij} \equiv x_{ij} $, with values
(energies) equal to the corresponding eigenvalues \cite{fyodorov2014topology,mehta2015energy}.

\begin{figure}[htp]
\begin{centering}
\includegraphics[width=\textwidth]{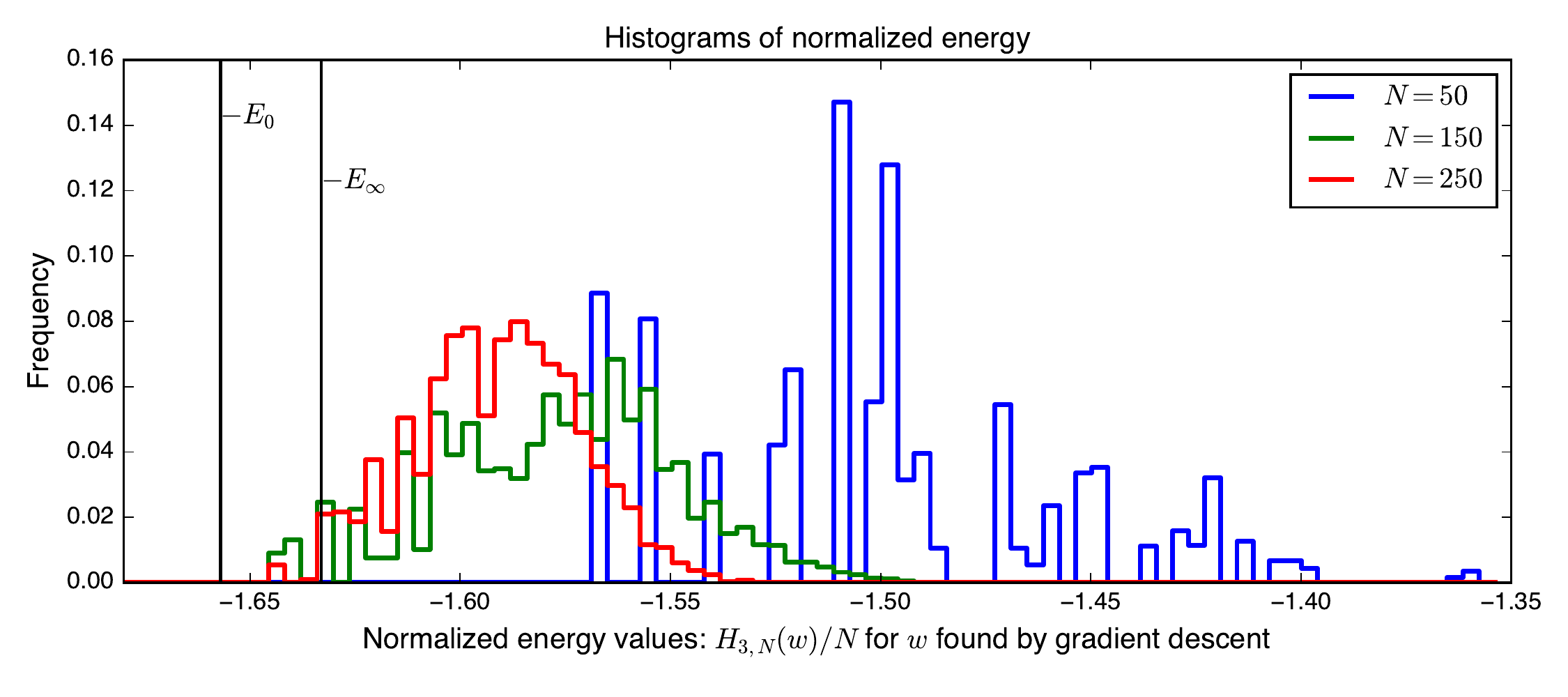}
\includegraphics[width=\textwidth]{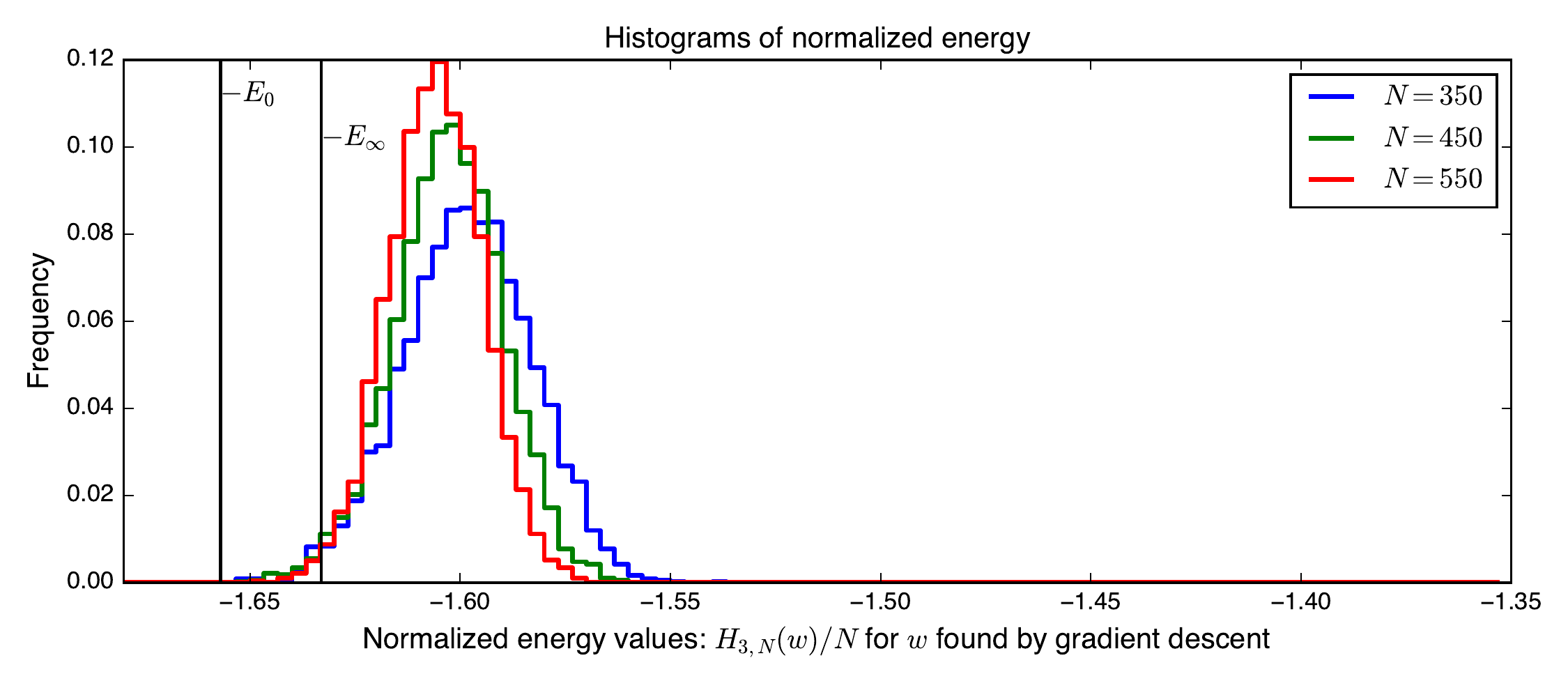}
\caption{Histogram for the energies of points found by gradient descent with
the Hamiltonian defined in Eq.~(\ref{eq:p3}), comparing low-dimensional and
high-dimensional systems. $-E_0$ denotes the ground state, and $-E_{\infty}$
denotes the large $N$ limit of the level for the bulk of the local minima.
\label{fig::spinglass_histogram}}
\end{centering}
\end{figure}

The picture is rather different when we look at polynomials with degree $p >
2$. When $p=3$ the polynomial becomes
\begin{equation}
H_{N, 3}(\vect{w})=\frac{1}{N}\sum_{i,j, k = 1}^N w_iw_jw_kx_{ijk},
\label{eq:p3}
\end{equation}
The normalisation factor $1/N$ for coupling coefficients
$x_{ijk} \sim \mathcal{N}(0,1)$, is chosen to make the extensive variables scale with $N$.
In other words, when
$\sum_{i=1}^{N} w_i^2 = N$, the variance of the Hamiltonian is proportional to
$N$. With this convenient choice of normalization, the results of Auffinger \textit{et al.}~\cite{auffinger2013random} show
that $H_{N, 3}(\vect{w})$ has exponentially many stationary points, including
exponentially many local minima (see \cite{mehta2013energy} for a complementary numerical study). 
In Fig.~\ref{fig::spinglass_histogram} we show
the distribution of minima for the normalised $H_{N, 3}(\vect{w})$ obtained by
gradient descent for various system sizes, $N$, and for a single realisation of the
coefficients. Since the variance of the
Hamiltonian scales with $N$, dividing by $N$ enables us to to compare energies for systems with different dimensions. 
The initial point is chosen uniformly at random from the sphere. The step
size is constant throughout the descent, until the norm of the
gradient becomes smaller than $10^{-6}$. For small $N$ the energies of the
minima are broadly scattered. However as the number of spins increases, the
distribution concentrates around a threshold. Further details of the
calculations can be found in Sagun \textit{et al.}~\cite{sagun2014explorations} 
(and \cite{Mehtaetal2017} for a complementary study).

To obtain a more precise picture for the energy levels of critical points, we will
focus on the level sets of the polynomial and count the number of critical
points in the interior of a given level set. Here, the polynomial is
assumed to be non-degenerate (its Hessian has nonzero eigenvalues). Then, a
critical point is defined by a point whose gradient is the zero vector, and a
critical point of index $k$ has exactly $k$
negative Hessian eigenvalues. Finally, our description of the number of
critical points will be in the exponential form and in the asymptotic, large $N$, limit.

Following Auffinger \textit{et al.}, let $A_u = \{{\bf w} \in \mathbb{R} ^N : H_{N,
3}({\bf w}) < u \text{ and } \sum_{i=1}^{N} w_i^2 = N \}$ be the set of points in the
domain of the function whose values lie below $u$. Also, let $C_k(A_u)$ denote
the number of stationary (critical) points with index $k$ in the set $A_u$.
Hence $C_k(A_u)$ counts the number of critical points with values
below $u$. Then, the main theorem of Ref.~\onlinecite{auffinger2013random}
produces the asymptotic expectation value $\mathbb{E}$ for the number of stationary points below
energy $u$:
\begin{equation}
    \lim_{N \rightarrow \infty} \frac{1}{N} \ln \mathbb{E}(C_k(A_u)) = \Theta_k(u)
    \label{eqn:complexity}
\end{equation}
where $\Theta$ is the complexity function, explicitly given in
Ref.~\onlinecite{auffinger2013random}. Note that the complexity function
$\Theta$ is non-decreasing and it is flat above some level. This result indicates
that there are no more finite index critical points at high levels, or to be more
precise, it is far less probable to find them. We denote this level as
$-E_{\infty}$. The second crucial quantity is when the complexity becomes
negative. This level has the property that there are no more critical
points of a specified index below it. We denote this level by $-E_k$, where
$k$ is the given index. For example, there are no more local minima below level
$-E_0$, which in turn means that the ground state is bounded from below by
$-E_0$. In particular, $\Theta$ approaches a constant for $ u > -E_{\infty} = 2
\sqrt{2/3} \approx -1.633$ and is bounded from below by $-E_0 \approx -1.657$.
We therefore have a lower bound for the value of the ground state, and all
stationary points exist in the energy band $-E_{0} \leq u \leq -E_{\infty}$.

\begin{figure}[htp]
\begin{centering}
\includegraphics[width=\textwidth]{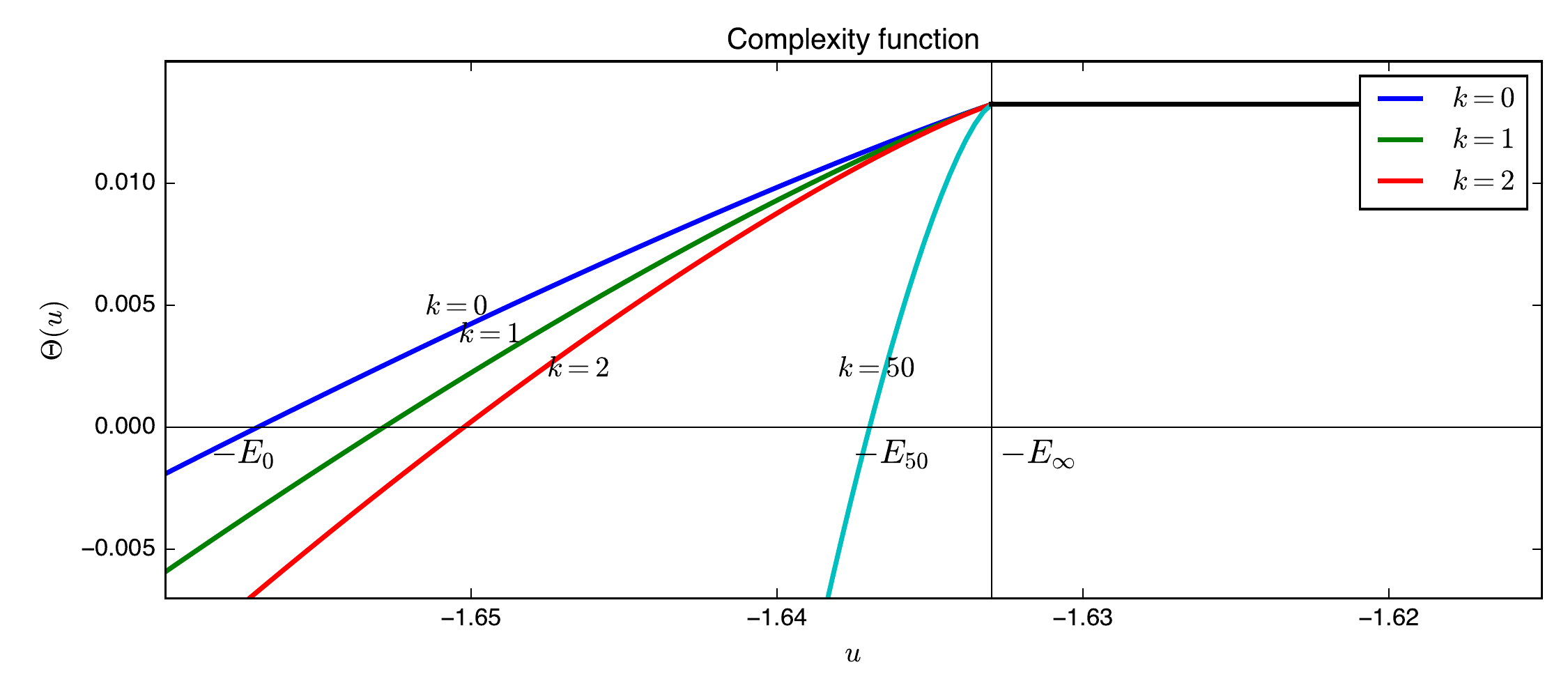}
\caption{Plots of the complexity function $\Theta_k$ 
defined in Eq.~\ref{eqn:complexity} for local minima,
and saddles of index 1, 2 and 50. In the band $(-E_0, -E_{50})$ there are only
critical points with indices $\{1,2,\dots,49\}$.}
\label{fig:complexity}
\end{centering}
\end{figure}

An inspection of the complexity function provides insight about the geometry at
the bottom of the energy landscape (Fig.~\ref{fig:complexity}). The ground
state is roughly at $u=-1.657$. For $u \geq -E_{\infty}$ we do not see any
local minima, because they all have values that are within the band $-E_{0}
\leq u \leq -E_{\infty}$. Moreover, in the same band the stationary points of
index $k > 0$ are outnumbered by local minima. In fact, this result holds
hierarchically for all critical points of finite index (recall that the result
is asymptotic so that by finite we mean fixing the index first and then taking
the limit $N \rightarrow \infty$). If we denote the $x$-axis
intercept of the corresponding complexity function $\Theta_k$ as $-E_k$, with
\begin{equation}
    \Theta_k(-E_k) = 0 \text{ for } k = 1, 2, \dots
    \label{eqn:intercept}
\end{equation}
Below the level $-E_k$ the function only has critical points of index strictly
less than $k$. This is consistent with the `glassiness' or `frustration'
that one would expect for such a system: a random quench is most likely to
locate a minimum around the $-E_{\infty}$ threshold and to find a lower energy
minimum numerous saddle points need to be overcome. This result suggests the following
scenario for finding local minima below the threshold. First identify an
initial local minimum through some minimisation algorithm. Since these points are
dominant at the $-E_{\infty}$ threshold, probabilistically speaking, the
algorithm will locate one around this value. Now we wish to jump over
saddle points to reach local minima with lower energies. Since the number of saddles
is much less than the number of local minima below the threshold, it may take
a lot longer to find them. This feature of the landscape could make finding the
global minimum a relatively difficult task. 
However, since basin-hopping \cite{lis87,lis88,WalesD97} removes downhill barriers, this approach might still be effective, depending on 
the organisation of the landscape.
Testing the performance of basin-hopping for such landscapes is an interesting
future research direction.
On the other hand, if the band $(-E_{0}, -E_{\infty})$ is narrow, which is the case for the spherical $3$-spin glass
Hamiltonian described above, then it may be sufficient to locate the first local
minimum and stop there, since further progress is unlikely.

\begin{figure}[htp]
\begin{centering}
\includegraphics[width=\textwidth]{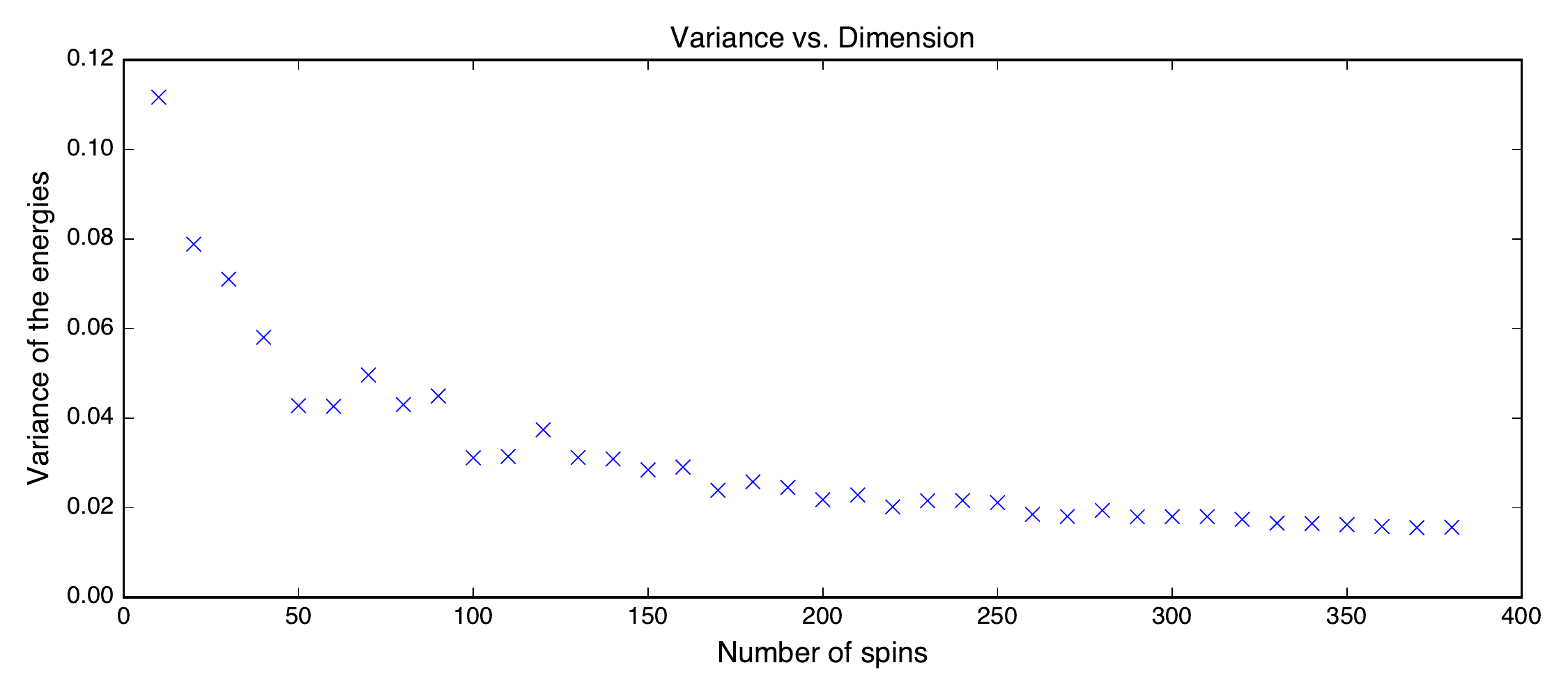}
\caption{Empirical variance of energies found by the gradient descent vs.~the number of spins of the Hamiltonian
defined in Eq.~(\ref{eq:p3}). \label{fig:shrinking_var}}
\end{centering}
\end{figure}

This scenario holds for the $p$-spin Hamiltonian with any $p \geq 3$ where
the threshold for the number of critical points is obtained
asymptotically in the limit $N \rightarrow \infty$. To demonstrate that it holds for
reasonably small $N$ Figure \ref{fig:shrinking_var} shows the results for
the $p=3$ case with increasing dimensions. The
concentration of local minima near the threshold increases rather quickly.

\begin{figure}
\centering
\begin{subfigure}[t]{0.32\textwidth}
\includegraphics[width=\textwidth]{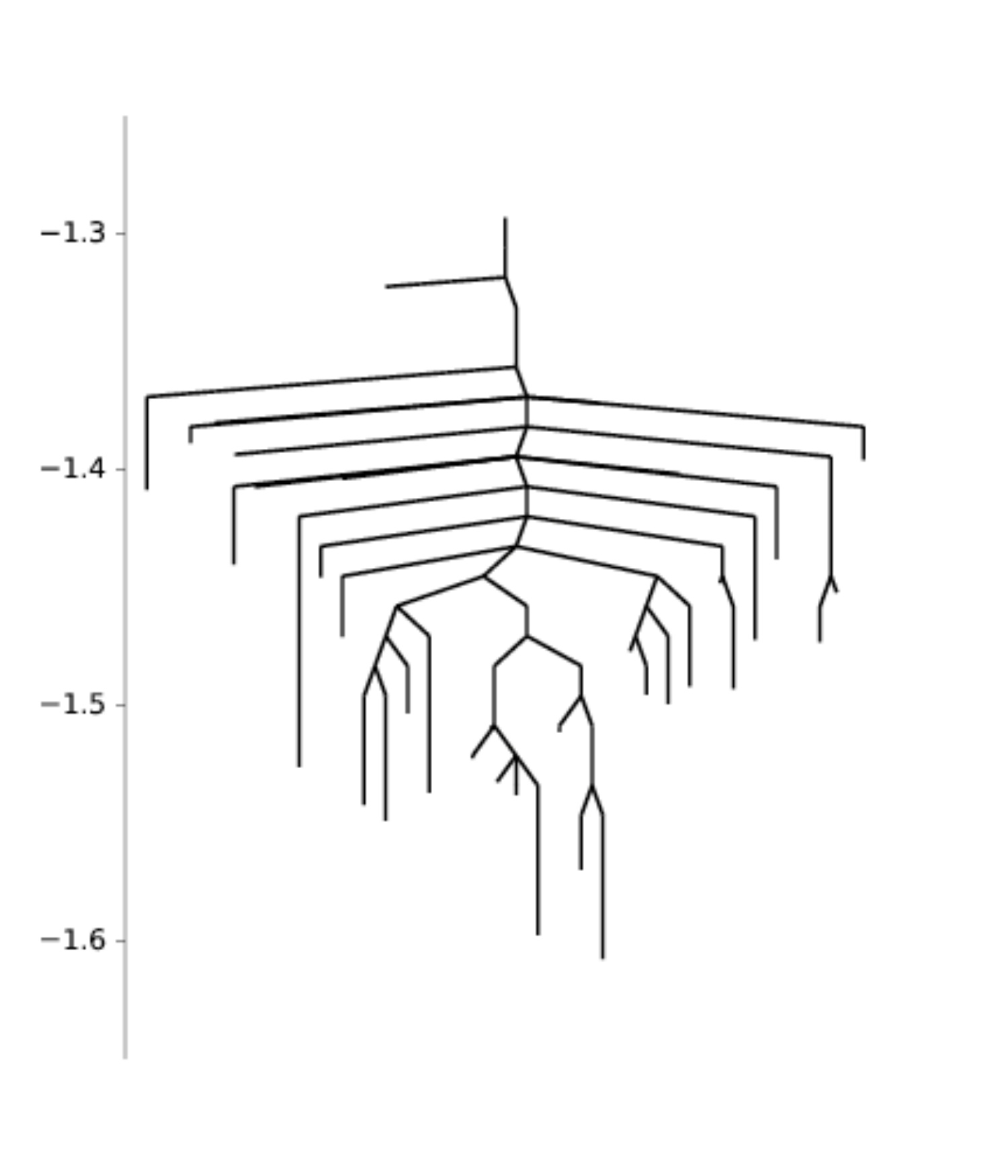}
\caption*{N=50}
\end{subfigure}
\begin{subfigure}[t]{0.32\textwidth}
\includegraphics[width=\textwidth]{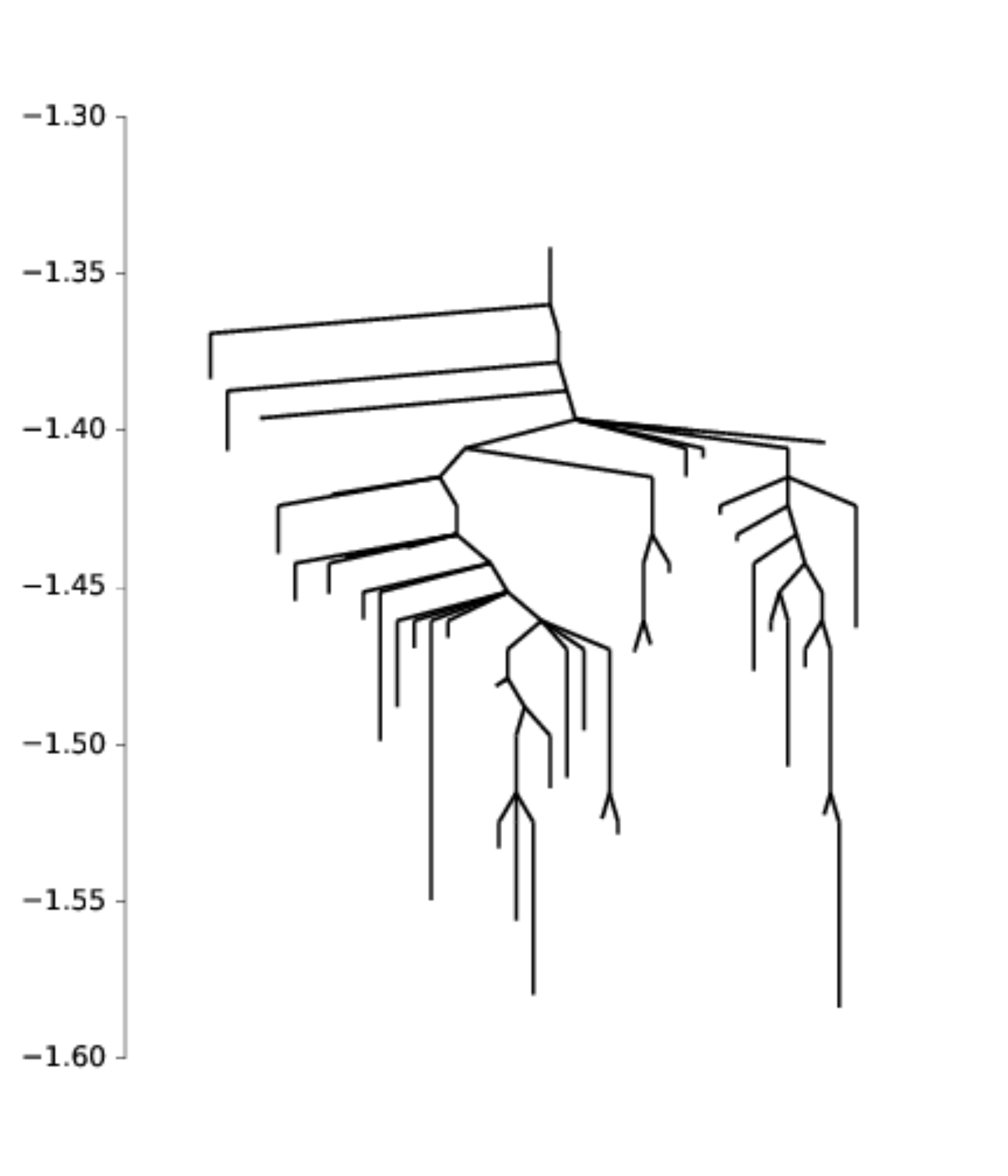}
\caption*{N=60}
\end{subfigure}
\begin{subfigure}[t]{0.32\textwidth}
\includegraphics[width=\textwidth]{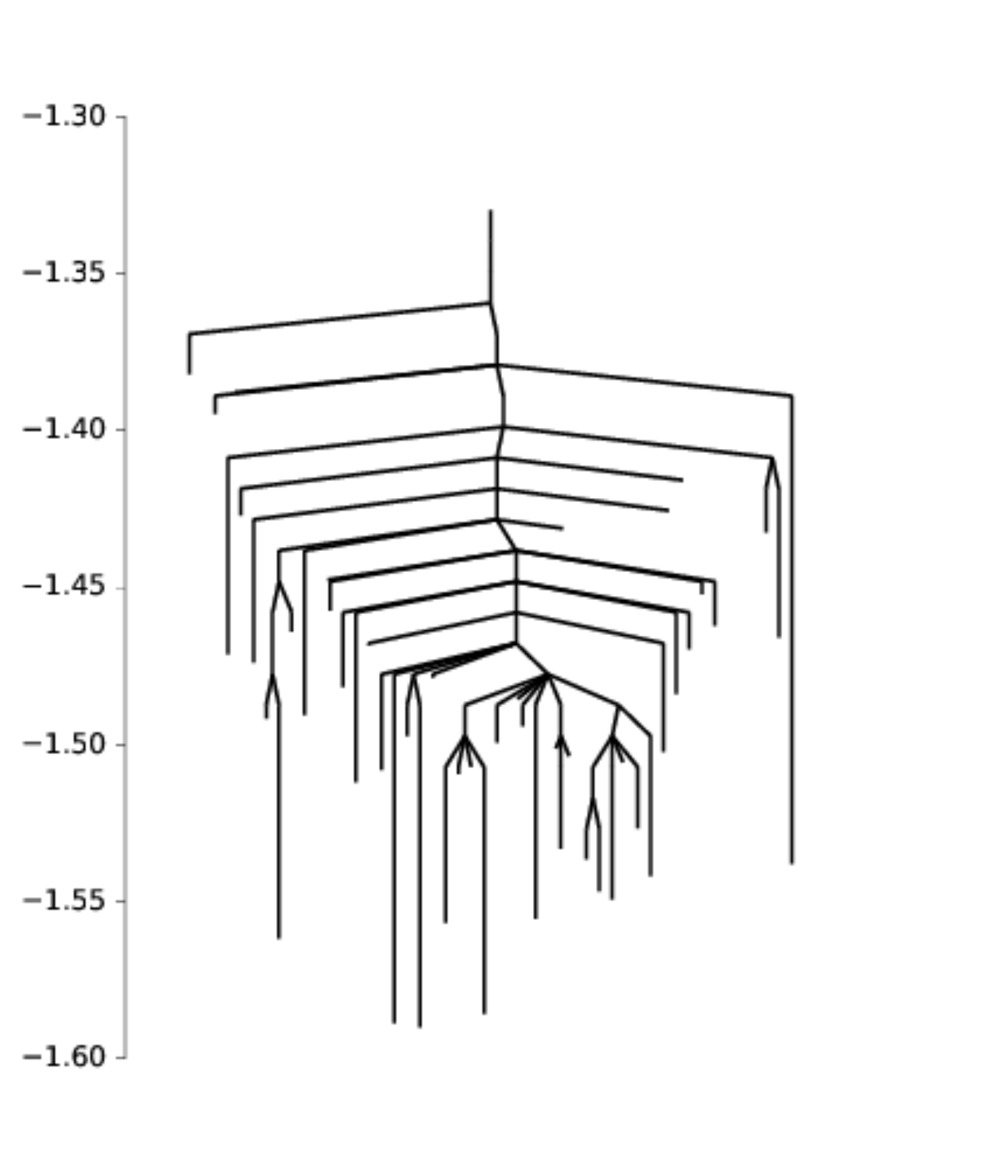}
\caption*{N=70}
\end{subfigure}
\begin{subfigure}[t]{0.32\textwidth}
\includegraphics[width=\textwidth]{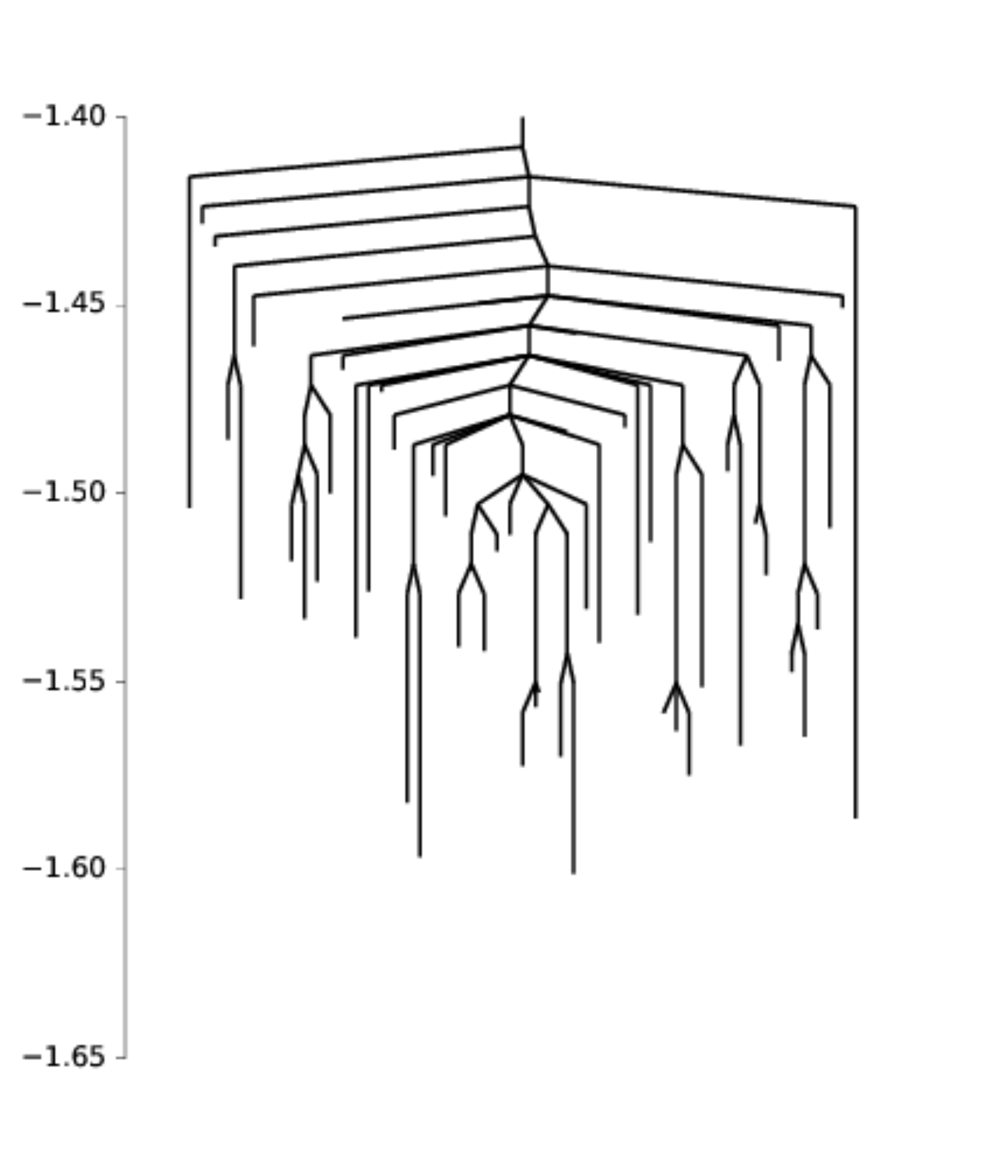}
\caption*{N=80}
\end{subfigure}
\begin{subfigure}[t]{0.32\textwidth}
\includegraphics[width=\textwidth]{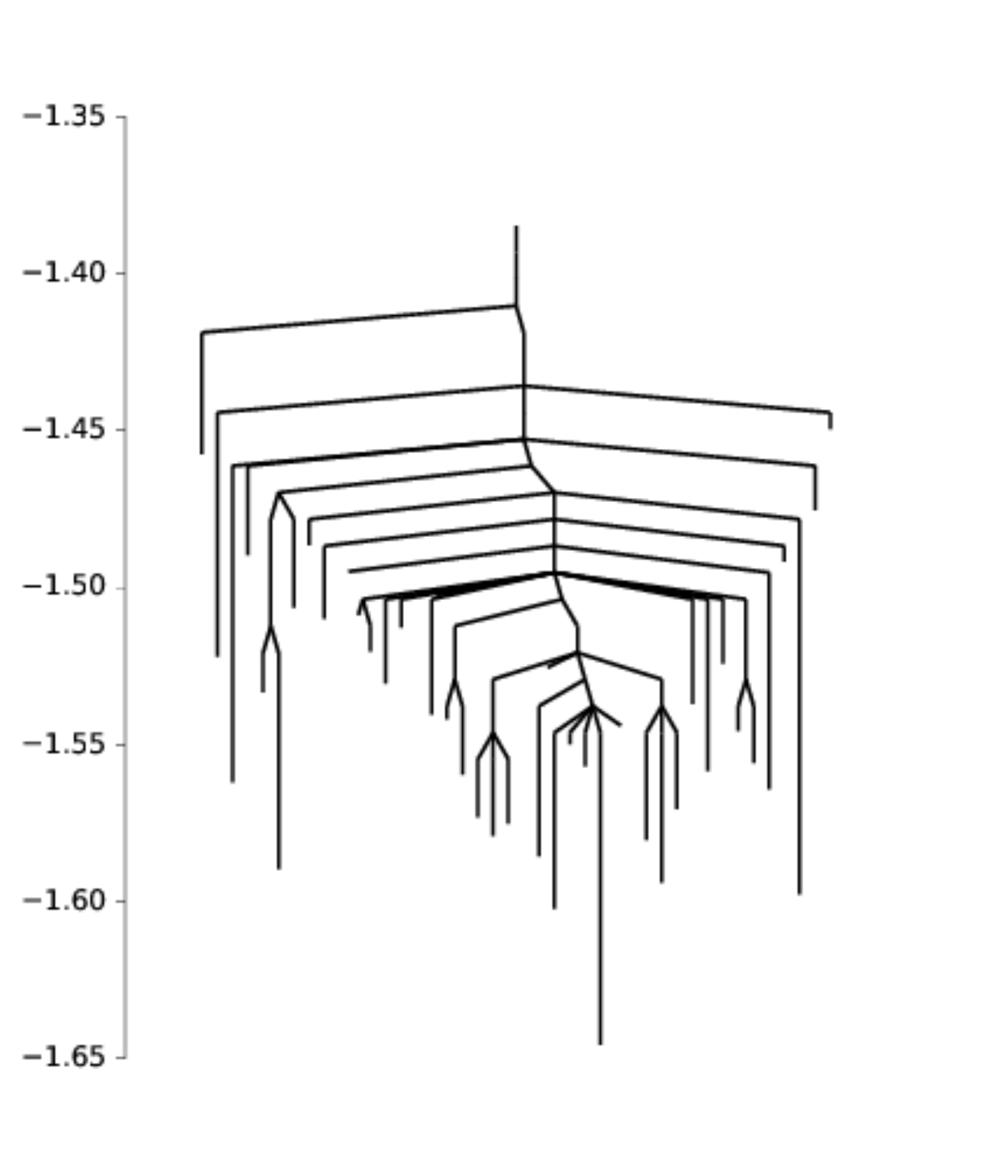}
\caption*{N=90}
\end{subfigure}
\begin{subfigure}[t]{0.32\textwidth}
\includegraphics[width=\textwidth]{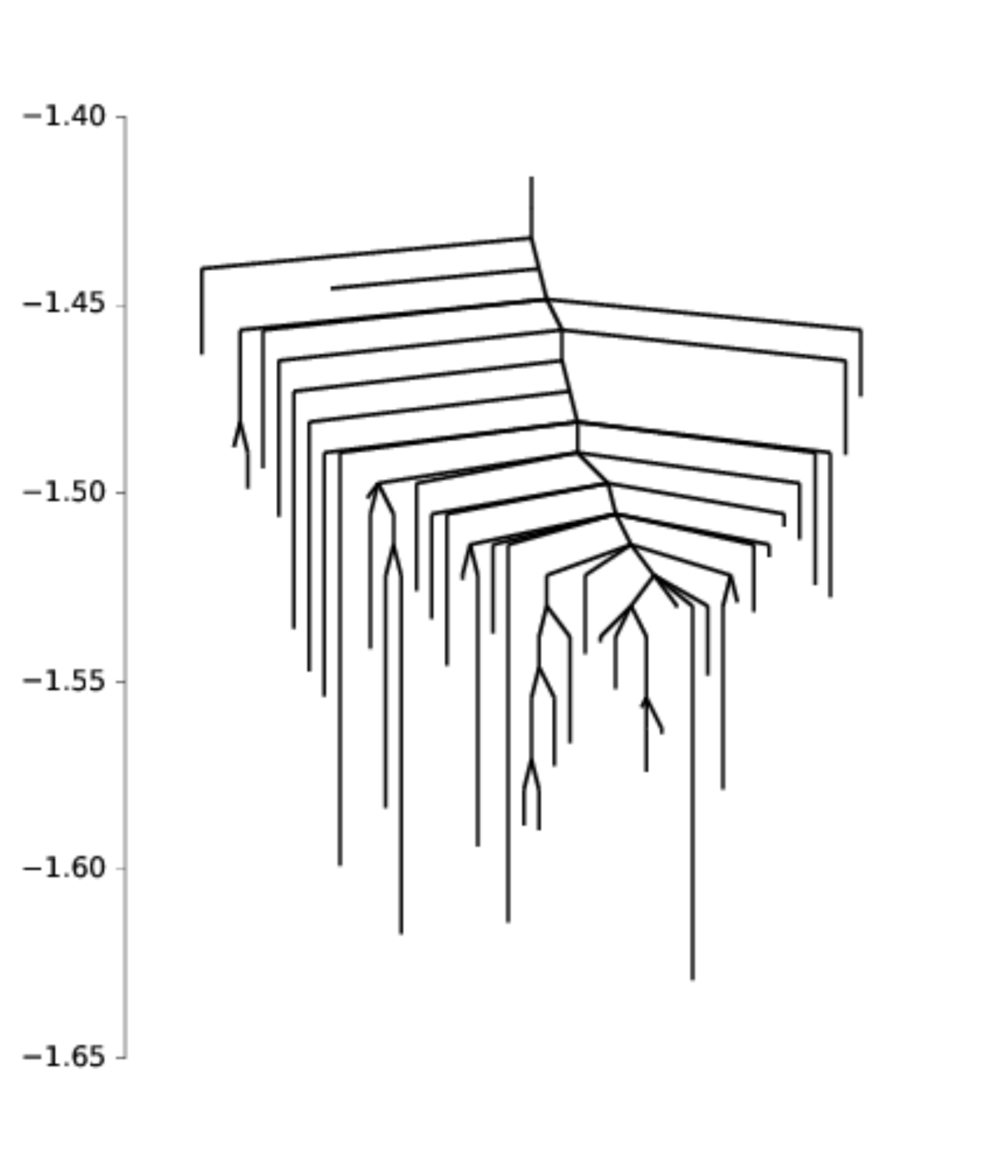}
\caption*{N=100}
\end{subfigure}
\caption{\label{fig::pspin_dg} Disconnectivity graphs for $p=3$ spin spherical
glass models of size $N \in [50,100]$. Each disconnectivity graph refers to a
single realisation of the coefficients $x_{ijk} \sim \mathcal{N}(0,1)$.
Frustration in the landscape is already visible for small system sizes, and the
minima appear to concentrate over a narrowing band of energy values for
as $N$ increases.
}
\end{figure}

In Fig.~\ref{fig::pspin_dg} we show disconnectivity graphs for $p=3$ spin
spherical glass models with sizes $N \in [50, 100]$ and fixed coefficients
$x_{ijk} \sim \mathcal{N}(0,1)$. The landscape appears to become more frustrated already
for small $N$, and the range over which local minima are found narrows with
increasing system size, in agreement with the concentration phenomenon
described in Ref.~\onlinecite{auffinger2013random}.

Interestingly, the concentration of stationary points phenomenon does not seem to be limited to
the \textit{p}-spin system. Related numerical
results\cite{sagun2014explorations} show that the same effect is
observed for a slightly modified version of the random polynomial, $\hat H_{N,
3}(\vect{w}^{(1)}, \vect{w}^{(2)}, \vect{w}^{(3)})= \sum_{i,j, k = 1}^N
w^{(1)}_i w^{(2)}_j w^{(3)}_k x_{ijk}$ defined on the three-fold product of
unit spheres. We do not yet have an analogue of the above theorem for $\hat H$,
so guidance from numerical results is particularly useful.

\subsection{Machine learning landscapes and concentration of stationary points}

The concept of complexity for a given function, as defined by 
the number and the nature of critical points on a given subset of the
domain, gives rise to a description of the landscape as outlined above. If the
energy landscape of machine learning problems is complex in this specific sense,
we expect to see similar concentration phenomena in the optimisation
of the corresponding loss functions. In fact, it is not straightforward to construct an
analogue of the $\theta$ function as in Eq.~(\ref{eqn:complexity}). However, we
can empirically check whether optimisation stalls at a level above the ground
state, as for the homogeneous polynomials with random coefficients described above.

\begin{figure}[htp]
\begin{centering}
\includegraphics[width=\textwidth]{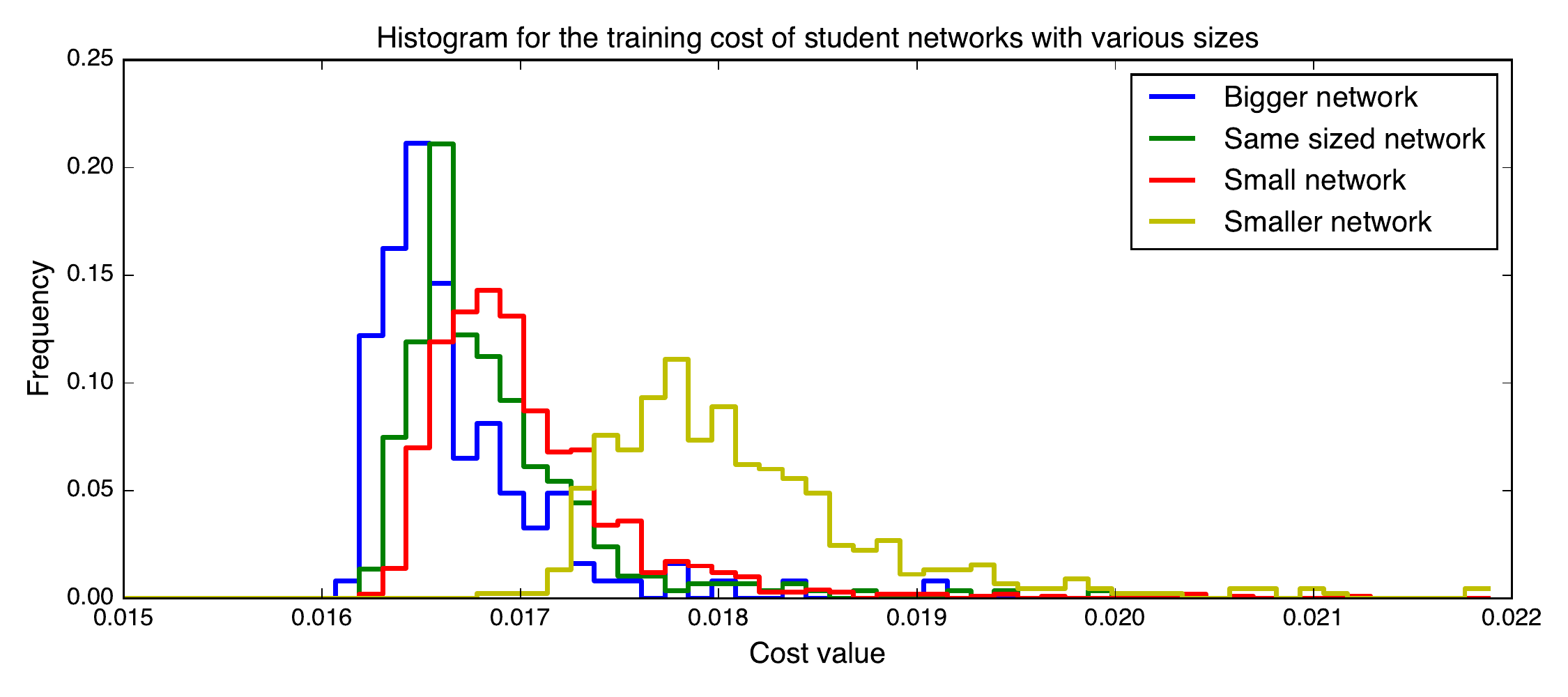}
\caption{Training on the student network of various sizes. The labels for the
network sizes are relative to the first network that is used to create new
labels. For the larger and equally sized networks the ground state is known to lie
at zero, yet the training stalls at a value around 0.016.}
\label{fig:teacher_student}
\end{centering}
\end{figure}

Let $D:={(\bm{x_1}, y_1), ..., (\bm{x_n}, y_n)}$ be $n$ data points with input $\bm{x_i}\in
\mathbb{R}^N$, and label $y \in \mathbb{R}$; and let $G(\bm{w}, \bm{x}) = \hat{y}$
describe the output that approximates the label $y$ parametrized by $\bm{w} \in
\mathbb{R}^M$. 
Further, let $\ell(G(\bm{w}, \bm{x}), y) = \ell(\hat{y}, y)$ be a
non-negative loss function. Once we fix the dataset, we can focus on the
averaged loss described by
\begin{equation}
    L(\bm{w}) = \sum_{i=1}^n \ell(G(\bm{w}, \bm{x_i}), y_i) 
    \label{eqn:loss_function}
\end{equation}
The function in Eq.~(\ref{eqn:loss_function}) is non-negative, but it is not
obvious where the ground state is located, and an empirical study could
be inconclusive. The following procedure fixes this problem. (1)
Create two identical models and split the training data in half. (2) Using the
first half of the data, train the first network, thereby obtaining a point
$w^*$ with a small value of $L(w^*)$ (3) Using $w^*$, create new labels for the
second half of the data, replacing the true labels with the output of the first
 model $G(\bm{w^*})$, (4) Using these new data pairs, $(\bm{x}, G(\bm{w^*}))$ train the second
network. This procedure ensures that the loss function for the second network
over the new dataset has configurations that have exactly value zero.
Simply by finding a copy of the first network, $\bm{w^*}$, the loss for the second
network will be $L(\bm{w^*}) = 0$. In fact, due to the permutation symmetry in the
parameters (see Figure \ref{fig:MLP}) the loss value remains zero for all the
points in the correct permutations of $\bm{w^*}$. Now the optimisation on the second
loss function has a known ground state at zero, and we can check empirically whether
optimisation stalls above that level \ref{fig:teacher_student} \cite{sagun2014explorations}.

We emphasise that the similarity between the $p$-spin Hamiltonian and the
machine learning loss function lies only in the concentration phenomena,
perhaps because the two functions share some underlying structure. It is likely that this
observation of concentration in the two systems, spin glasses and deep
learning, are due to different reasons that are peculiar to their
organisation. It is also possible that such concentration
behaviour is universal, in the sense that it can be observed in various
non-convex and high dimensional systems for which the two systems above are
just examples. However, if we wish to describe the ML landscape in terms of glassy behaviour,
we might seek justification through the non-linearities
in neural networks (the hidden layer in Figure \ref{fig:MLP}). In some sense,
the non-linearity of a neural network is the key that introduces competing
forces in which neurons can be inhibitory or exhibitory. This behaviour may
introduce a similar structure to the signs of interaction coefficients in spins,
introducing frustration. We note that these interpretations are rather
speculative at present. Another problem that
requires further research is identification of all critical points, not only
the ones with low index. A more systematic way to identify the notion of
complexity is through finding all of the critical points of a given function.
A further challenge lies in the degeneracy of the systems in deep learning.
The random polynomials that we considered have non-degenerate Hessian eigenvalue
spectra for stationary points at the bottom
of the landscape. However, a typical neural network has most
Hessian eigenvalues near zero \cite{sagun2016singularity}.

Recently, a complementary study of the minima and saddle points of the $p$-spin model 
has been initiated using the numerical polynomial homotopy continuation method \cite{sommese2005numerical,mehta2011finding},
which guarantees to find \textit{all} the minima, transition states and higher index saddles for
moderate sized systems \cite{mehta2015energy,mehta2013energy}. An advantage of this approach is that one can also analyse
the variances of the number of minima, transition states, etc., providing further insight into the landscape of the $p$-spin model.
A study for larger values of $p$, analysing the interpretation in terms of a deep learning network, 
is in progress \cite{Mehtaetal2017}.

\section{Basin volume calculations: quantifying the energy landscape}

The enumeration of stationary points in the energy landscape provides a direct
measure of complexity. This quantity is directly related to 
the landscape
entropy \cite{SciortinoKT00,BogdanWC06,MengABM10,Wales10b} (to be distinguished from
the well entropy associated with the vibrational modes
of each local minimum\cite{berthier2014novel}) and is crucial for
understanding the emergent dynamics and thermodynamics. In this context other
important questions include determining the level of the stationary points (as
we discussed in Sec.~\ref{sec::pspin}) and the volume of their basins of
attraction. These volumes are of great practical interest because they provide an
\textit{a priori} measure of the probability of finding a particular minimum
following energy minimisation. This probability is particularly important within the
context of non-convex optimisation problems and machine learning, where an
established protocol to quantify the landscape and the \textit{a priori}
outcome of learning is lacking.

The development of general methods for enumerating the number of
accessible microstates in a system, and ideally their individual probabilities,
is therefore of great general interest. As discussed in Sec.~\ref{sec::pspin}, for a few
specific cases there exist methods -- either analytical or numerical -- capable
of producing exact estimates of these numbers. However, these techniques are
either not sufficiently general or simply not practical for
problems of interest. To date, at least two general and practical
computational approaches have been developed. `Basin-sampling'  
\cite{Wales13} employs a model anharmonic density of states for local minima
organised in bins of potential energy,
and has been applied to atomic clusters, including benchmark systems with double
funnel landscapes that pose challenging issues for sampling. The
mean basin volume (MBV) method developed by Frenkel and co-workers
\cite{xu2011direct, asenjo2014numerical, MartinianiSSWF16,
martiniani2016structural} is similar in spirit to basin-sampling, 
but is based on thermodynamic integration and, being
completely general, requires no assumptions on the shape of the basins
(although thus far all examples are limited to the enumeration of the minima).
MBV has been applied in the context of soft sphere packings and has facilitated
the direct computation of the entropy in two \cite{xu2011direct,
asenjo2014numerical} and three \cite{MartinianiSSWF16} dimensions.
Furthermore, the technique has allowed for a direct test of the Edwards
conjecture in two dimensions \cite{martiniani2016some}, suggesting that only at
unjamming -- when the system changes from a fluid to a solid, which is the density
of practical significance for many granular systems -- the entropy is maximal
and all packings are equally probable.

Despite the high computational cost, the MBV underlying principle is
straightforward. Assuming that the total configuration volume $\mathcal{V}$ of the energy
landscape is known (simply $\mathcal{V} = V^N$ for an ideal gas of interacting
atoms), if we can estimate the mean basin volume of all states, the number of
minima is simply
\begin{equation}
\Omega = \frac{\mathcal{V}}{\langle v_{\text{basin}} \rangle},
\end{equation}  
where $\langle v_{\text{basin}} \rangle$ is the unbiased average volume of the
basins of attraction. We distinguish the biased from the unbiased distribution
of basin volumes because, when generating the minima following minimisation from
uniformly sampled points in $\mathcal{V}$, they will be sampled in proportion
to the volume of the basin of attraction, and therefore the observed
distribution of $v_{\text{basin}}$ is biased. A detailed discussion of the
unbiasing procedure for jammed soft-sphere packings is given in
Ref.~\onlinecite{MartinianiSSWF16}. The Boltzmann-like entropy of the system is
then simply $S_B = \ln \Omega- \ln N! $. Similarly, from knowledge of the biased (observed)
distribution of basin volumes $v_i$ alone, one can compute the Gibbs-like (or
Shannon) entropy $S_G = - \sum_{i=1}^\Omega p_i \ln p_i- \ln N! $, where $p_i = v_i/\mathcal{V}$
is the relative probability for minimum $i$.

The computation of the basin volume is performed by thermodynamic integration. In essence, we perform a series of Markov chain Monte Carlo random walks within the basin applying different biases to the walkers and, from the distributions of displacements from a reference point (usually the minimum), compute the dimensionless free energy difference between a region of known volume and that of an unbiased walker. In other words
\begin{equation}
f_{\text{basin}} = f_{\text{ref}} + (\hat{f}_{\text{basin}}-\hat{f}_{\text{ref}})
\end{equation}
where the dimensionless free energy is $f = - \ln v$ and the hat refers to quantities estimated up to an additive constant by the free energy estimator of choice, either Frenkel-Ladd \cite{frenkel1984new, MartinianiSSWF16} or the multi-state Bennet acceptance ratio method (MBAR) \cite{shirts2008statistically, martiniani2016structural}. The high computational cost of these calculations is due to the fact that in order to perform a random walk in the body of the basin, a full energy minimisation is required to check whether the walker has overstepped the basin boundary.

Recently the approach has been validated when the dynamics determining the
basin of attraction are stochastic in nature \cite{frenkel2016monte}, which is
precisely the situation encountered in the training by stochastic optimisation
of neural networks and other non-convex machine learning problems. The
extension of these techniques to machine learning is another exciting prospect, as it
would provide a general protocol for quantifying the machine learning landscape
and establishing, for instance, whether the different solutions to learning
occur with different probabilities and, if so, what their distribution is.
This characterisation of the learning problem could help to develop
better models, as well as better training algorithms.

\section{Conclusions}

In this Perspective we have applied theory and computational techniques from
the potential energy landscapes field \cite{Wales03} to analyse problems in machine learning. 
The multiple solutions that can result from optimising fitting functions to
training data define a machine learning landscape \cite{BallardSDW16}, where the cost function 
that is minimised in training takes the place of the molecular potential energy function.
This machine learning landscape can be explored and visualised using methodology
transferred directly from the potential energy landscape framework.
We have illustrated how this approach can be used through examples taken from
recent work, including analogies with thermodynamic properties, such as the
heat capacity, which reports on the structure of the equilibrium properties of the
solution space as a function of a fictitious temperature parameter.
The interpretation of ML landscapes in terms of analogues of molecular structure
and transition rates is an intriguing target for future work.

Energy landscape methods may
provide a novel way of addressing one of the most intriguing 
questions in the machine learning research, namely \textit{why does machine learning work so well?} One way to ask this 
question more quantitatively is to investigate why we can usually find a good candidate for the global minimum 
of a machine learning cost function relatively quickly,
even in the presence of so many local minima. The present results
suggest that the landscape for a range of models are single-funnel-like, i.e.~the largest basin 
of attraction is that of the global minimum, and the downhill barriers that separate it from local minima are 
relatively small.
This organisation facilitates
rapid relaxation to the global minimum for global optimisation techniques,
such as basin-hopping. Another possible explanation is that
many local minima provide fits that are competitive with
the global minimum~\cite{Dauphin2014, Bray2007, sagun2014explorations}. 
In fact, these two scenarios are also compatible, so that global optimisation
leads us downhill on the landscape, where we encounter local minima 
that provide predictions or classifications of useful accuracy.

The ambition to develop 
more fundamental connections between machine
learning disciplines and computational chemical physics could be very productive. 
For example, there have recently been many physics-inspired contributions 
in machine learning, including thermodynamics-based models 
for rational decision-making \cite{Ortega2012}, 
generative models from non-equilibrium simulations \cite{Dickstein2015}. The hope is that such connections can provide better intuition about the machine
learning problems in question, and perhaps also the underlying physical theories
used to understand them.

\section{Acknowledgements}
It is a pleasure to acknowledge discussions with Prof.~Daan Frenkel, Dr Victor Ruehle, Dr Peter Wirnsberger, Prof.~G\'erard Ben Arous, and Prof.~Yann Lecun. 
This research was funded by EPSRC grant EP/I001352/1, the Gates Cambridge Trust, and the ERC. 
DM was in the Department of Applied and Computational 
Mathematics and Statistics when this work was performed, and his current affiliation is Department of Systems, United Technologies Research 
Center, East Hartford, CT, USA.

%

\end{document}